\documentclass{article}

\usepackage[final]{neurips_2025}
\usepackage{graphicx}

\usepackage[utf8]{inputenc} 
\usepackage[T1]{fontenc}    
\usepackage{hyperref}       
\usepackage{url}            
\usepackage{booktabs}       
\usepackage{amsfonts}       
\usepackage{amsmath}
\usepackage{nicefrac}       
\usepackage{microtype}      
\usepackage{xcolor}         
\usepackage{cleveref}
\usepackage{wrapfig,algorithm,algorithmic}

\usepackage{hhline}
\usepackage{subcaption}
\usepackage{enumitem}
\usepackage{float} 
\usepackage{wrapfig}
\usepackage{amsthm}
\theoremstyle{plain}
\newtheorem{theorem}{Theorem}[section]

\theoremstyle{definition}

\theoremstyle{remark}

\newif\ifdraft
\drafttrue

\usepackage{todonotes}

\newcommand{\R}[0]{\mathbb{R}}
\newcommand{\C}[0]{\mathbb{C}}

\newcommand{\logistic}[1]{\sigma_{#1}}
\newcommand{\charfun}[1]{\varphi_{#1}}
\newcommand{\empcharfun}[1]{\hat{\varphi}_{#1}}
\newcommand{\measure}[1]{\mathbb{P}_{#1}}
\newcommand{\component}[1]{_{\left[ #1 \right]}}
\newcommand{\expectation}[2][]{\mathbb{E}_{#1}\left[ #2 \right]}
\newcommand{\norm}[1]{\left\lVert#1\right\rVert}
\newcommand{\absval}[1]{\left\lvert#1\right\rvert}

\usepackage{acronym}

\usepackage{titlesec}
\titlespacing*{\paragraph}{0pt}{2pt}{1em}

\setlist[itemize,enumerate]{topsep=0pt, itemsep=0pt, leftmargin=15pt}

\title{
Simple and Effective \\ Specialized Representations for Fair Classifiers
}

\author{%
  Alberto~Sinigaglia\thanks{These authors contributed equally.} \\
  Human Inspired Technology Research Center\\
  University of Padua\\
  \texttt{alberto.sinigaglia@phd.unipd.it} \\
  \And
  Davide~Sartor\footnotemark[1] \\
  Department of Information Engineering\\
  University of Padua\\
  \texttt{davide.sartor.4@phd.unipd.it} \\
  \And
  Marina~Ceccon \\
  Department of Information Engineering\\
  University of Padua\\
  \texttt{marina.ceccon@phd.unipd.it} \\
  \And
  Gian~Antonio~Susto\\
  Department of Information Engineering\\
  University of Padua\\
  \texttt{gianantonio.susto@unipd.it} \\
}

\newacro{MLP}{Multi Layer Perceptron}
\newacro{CF}{Characteristic Function}
\newacro{CFD}{Characteristic Function Distance}
\newacro{LR}{Logistic Regression}

\begin{document}

\maketitle

\begin{abstract}
Fair classification is a critical challenge that has gained increasing importance due to international regulations and its growing use in high-stakes decision-making settings.
Existing methods often rely on adversarial learning or distribution matching across sensitive groups; however, adversarial learning can be unstable, and distribution matching can be computationally intensive.
To address these limitations, we propose a novel approach based on the characteristic function distance. Our method ensures that the learned representation contains minimal sensitive information while maintaining high effectiveness for downstream tasks. 
By utilizing characteristic functions, we achieve a more stable and efficient solution compared to traditional methods.
Additionally, we introduce a simple relaxation of the objective function that guarantees fairness in common classification models with no performance degradation. 
Experimental results on benchmark datasets demonstrate that our approach consistently matches or achieves better fairness and predictive accuracy than existing methods. 
Moreover, our method maintains robustness and computational efficiency, making it a practical solution for real-world applications.
\end{abstract}

\section{Introduction}
Algorithmic fairness has become a central concern in deploying automated decision-making systems, especially in high-stakes domains like hiring, lending, criminal justice, and healthcare \cite{blake2024algorithmic, wang2022brief, gillis2024operationalizing, fabris2024fairness}. The growing reliance on these systems has raised concerns about their potential to reinforce or amplify societal biases \cite{barocas-hardt-narayanan, dualgorithmic, vetro2021data, brzezinski2024properties}. In response, a large body of research has focused on detecting, analyzing, and mitigating bias throughout the algorithmic pipeline \cite{siddique2023survey, zemel2013learning, mcnamara2019costs, cooper2024arbitrariness}.

Numerous fairness definitions and metrics have been proposed, reflecting a wide range of normative and technical perspectives \cite{dwork2012fairness, hardt2016equality, du2020fairness}. A common criterion is statistical independence, which requires that model predictions be independent of sensitive attributes like ethnicity or gender \cite{R_z_2021, barocas-hardt-narayanan}. 
A simple strategy for enforcing this criterion is to exclude the sensitive attribute from the model's input features, an approach commonly referred to as \textit{Fairness through Unawareness} \cite{cornacchia2023auditing}.
However, this method is often ineffective in practice, as sensitive information may still be indirectly captured through other correlated features, so-called proxy variables \cite{pitoura2022fairness}. 
Removing all proxies typically results in significant degradation of performance, as task-relevant information is also discarded \cite{cornacchia2023auditing}.

To better balance fairness and predictive accuracy, many approaches aim to learn fair representations by projecting input data into a latent space that obscures sensitive information while preserving task-relevant structure \cite{zemel2013learning, madras2018learning, moyer2018invariant, balunovic2022fair}. Once optimized, these representations can replace the original data during both training and testing, enabling fair decision-making in scenarios where sensitive attributes must not be explicitly used, such as determining loan approvals or hiring decisions.
These approaches typically fall into two categories, which we refer to as \textit{specialized} and \textit{general}. Specialized representations are tailored to a specific task, filtering out sensitive and irrelevant features for improved performance. 
General representations, in contrast, are task-agnostic and aim for broad applicability across multiple tasks while being insensitive to protected attributes, but are harder to learn and may compromise predictive accuracy. Returning to the loan approval example, specialized representations learned in that context would be suitable only for predicting creditworthiness. If reused for another task, e.g. default risk or customer lifetime value, they might lack task-relevant information that was discarded during the fairness optimization. General representations, instead, preserve information that remains broadly useful across prediction tasks while still mitigating sensitivity to protected attributes.

A popular class of methods uses Variational Autoencoders (VAEs) to learn fair, specialized representations \cite{louizos2017variationalfairautoencoder, moyer2018invariant}. In VAE-based approaches, fairness is optimized indirectly, either via a variational bound or Maximum Mean Discrepancy (MMD), which can misalign with true fairness goals and yield suboptimal results \cite{balunovic2022fair}. 

\begin{figure}[t!]
    \centering
    \includegraphics[width=\textwidth]{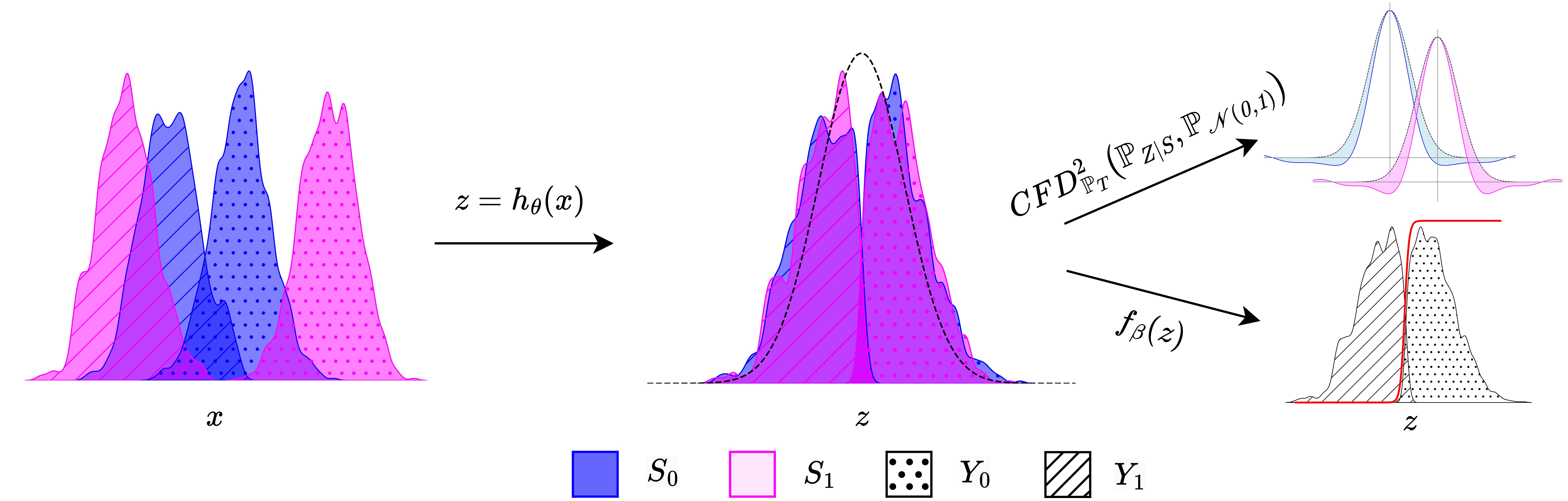}
    \caption{
        Overview of the proposed approach.
        Each sample $X$ is associated with a sensitive attribute $S$ and a target label $Y$. 
        The conditional distribution $ \measure{X|S} $ is mapped to a new distribution $\measure{Z \mid S}$, 
        which is encouraged to resemble a Gaussian Distribution via the \ac{CFD}. The encoded representation $Z$ minimizes $\Delta(\measure{Z \mid S_0}, \measure{Z \mid S_1})$ while retaining task-relevant information.
    }
    \vskip -0.1cm
    \label{fig:frontpage}
\end{figure}

Other methods rely on Adversarial Networks that ensure specialized fair representations through a minimax game between a predictor and a fairness discriminator \cite{edwards2015censoring, jaiswal2020invariant}. 
Such approaches are often unstable and sensitive to hyperparameters, making the fairness-performance trade-off more challenging to manage \cite{elazar2018adversarial}. 
Additionally, fairness guarantees are tied to the specific adversary used during training and may fail if a stronger or different adversary is applied later \cite{moyer2018invariant, xu2020theory, gupta2021controllable, song2019overlearning}.

Fair Normalizing Flows (FNF) \cite{balunovic2022fair} offer representations that are robust to any adversary, positioning itself as a state-of-the-art alternative to adversarial methods. 
Nonetheless, it has key limitations. 
It requires sensitive attribute information at inference, conflicting with privacy goals. 
It also trains a separate model for each sensitive group, which is computationally expensive, especially with multi-class attributes or large datasets. 
Crucially, while FNF aims to learn general representations, the inclusion of a classification loss can bias the model toward task-specific features.
This may introduce group-specific information, potentially weakening fairness guarantees through representation bias.

There is an inherent trade-off between general and specialized representations: while general representations are broadly useful and transferable, achieving high predictive performance often requires some degree of specialization. Even models like FNF, which explicitly aim for generalization, ultimately rely on task-specific loss functions that guide representations toward specialization to optimize accuracy. This reliance suggests that specialization is not merely a byproduct but a necessary condition for strong predictive performance. Recognizing this, it becomes clear that embracing specialization is not only pragmatic but essential.
Adversarial approaches naturally embody this perspective by tailoring representations to specific prediction tasks. However, they come with significant limitations, most notably, weaker guarantees around fairness and robustness \cite{balunovic2022fair, moyer2018invariant, song2019overlearning}. This underscores the need for alternative methods that, like adversarial techniques, pursue specialization, but do so through more transparent and controllable means, offering stronger fairness guarantees as a result.

To address these challenges, we propose a novel framework that embraces specialization in representations while rigorously enforcing fairness, without relying on adversarial training.
In addition to stronger guarantees, our method is significantly simpler and more lightweight than existing approaches. This further justifies the aim to achieve specialized representations, as the advantages of transferability offered by general representations might not outweigh the benefit of higher prediction accuracy in settings where re-training for a different task is computationally lightweight.
Our key contributions are summarized as follows:
\begin{itemize}
    \item We introduce a new approach based on distribution matching via characteristic functions.
    This formulation avoids reliance on auxiliary components such as variational autoencoders, adversarial networks, or normalizing flows.
    \item We derive a simplified version of the proposed framework tailored to classification, which allows for formal guarantees on the sensitive information accessible to downstream classifiers.
    \item We conduct extensive experiments demonstrating that our method effectively removes sensitive information while matching SotA accuracy and delivering significantly fairer representations.
\end{itemize}
In summary, our approach offers a principled alternative for learning fair, specialized representations. It seamlessly combines simplicity, training stability, and robust fairness guarantees, all without relying on sensitive attributes during inference. 
This makes our method both practical and privacy-conscious, showing promising improvements over the current state-of-the-art \cite{balunovic2022fair}.

\section{Related Work}
\paragraph{Variational Autoncoders} In the context of fairness through representation learning, prior work has explored the use of Variational Autoencoders (VAEs) to disentangle sensitive attributes from learned data representations \cite{louizos2017variationalfairautoencoder, moyer2018invariant, creager2019flexibly, locatello2019fairness}. 
The encoder extracts task-relevant representations, while the decoder reconstructs the input. Though reconstruction may retain extra information, the optimization encourages representations tailored to prediction. 
A notable approach is the Variational Fair Autoencoder (VFAE) \cite{louizos2017variationalfairautoencoder}, which extends the standard VAE framework to enforce invariance in the latent space with respect to protected attributes. 
VFAE achieves this by introducing a penalty that explicitly encourages independence between the latent representation and the sensitive attributes. 
A related method was proposed by Moyer et al.~\cite{moyer2018invariant}, who combine ideas from VAEs and the Variational Information Bottleneck (VIB) to learn representations that are both informative and robust to variations in sensitive inputs.

\paragraph{Adversarial Learning} 
Another line of research focuses on fairness-aware learning through adversarial and information-theoretic methods to induce invariant representations \cite{xie2017controllable, roy2019mitigating, song2019learning}. 
In adversarial settings, a model such as an encoder or predictor is trained in opposition to an adversary whose goal is to recover sensitive attributes from the learned representation. This adversarial pressure encourages the model to generate latent features that do not reveal sensitive information, thereby promoting invariance with respect to protected variables \cite{edwards2015censoring, feng2019learning}.
Xie et al.~\cite{xie2017controllable} proposed a general adversarial framework for learning representations that are invariant to arbitrary nuisance attributes. 
Their method formulates the learning process as a three-player minimax game involving an encoder, a task-specific predictor, and a discriminator that attempts to infer the nuisance attribute. 
Madras et al.~\cite{madras2018learning} introduced Learning Adversarially Fair and Transferable Representations (LAFTR), which incorporates adversarial objectives aligned with specific fairness definitions, to learn representations that remain fair even when deployed by downstream classifiers without explicit fairness constraints.
Roy and Bodetti \cite{roy2019mitigating} extended this adversarial paradigm by proposing MaxEnt-ARL, which maximizes the entropy of the adversary's prediction of the sensitive attribute rather than minimizing its accuracy. This approach improves privacy, as it offers the practical benefit of not requiring access to sensitive labels during encoder training. 
Jaiswal et al.~\cite{jaiswal2020invariant} introduced Adversarial Forgetting, a framework that decouples the learning of rich representations from the selective forgetting of sensitive or nuisance information via a dedicated forget-gate mechanism. 
Building on adversarial fairness frameworks, FR-Train \cite{roh2020fr} incorporates a mutual information-based formulation and adds a second adversary to enhance robustness to poisoned data.

\paragraph{Other Approaches} 
Recognizing the limitations of adversarial frameworks, particularly their instability and lack of formal guarantees \cite{elazar2018adversarial, gupta2021controllable, feng2019learning}, researchers have explored alternative paradigms for learning fair representations. 
Jiang et al.~\cite{jiang2020wasserstein} propose a theoretically grounded approach that enforces demographic parity by minimizing the Wasserstein-1 distance between model outputs across different sensitive groups.
Tucker and Shah~\cite{tucker2022prototype} present Concept Subspace Networks (CSNs), a prototype-based architecture that unifies fair and hierarchical classification within a single model. 
Ultimately, building on these ideas, Balunović et al.~\cite{balunovic2022fair} introduce Fair Normalizing Flows (FNF), a framework that provides provable fairness guarantees against any downstream adversary. 
FNF represents the current state-of-the-art in terms of fair representations. It employs separate normalizing flow encoders for each sensitive group and ensures fairness by minimizing the statistical distance between the resulting latent distributions. 
This formulation enables exact likelihood computation in the latent space, allowing for theoretical upper bounds on unfairness for any downstream classifier, a property not commonly achieved in existing fair representation learning methods. Furthermore, Balunović et al.~\cite{balunovic2022fair} show how adversarial learning inherently causes a false sense of fairness \cite{feng2019learning, xu2020theory, elazar2018adversarial, gupta2021controllable}. Indeed, they show how multiple methods based on such techniques share a tendency to break once the learned representations are tested on more powerful families of classifiers. Though adversarial learning is the only approach offering specialized representations, such evidence reinforces the need to develop novel directions to achieve them without the instabilities of adversarial learning.

\section{Background}
\label{sec:background}
Let $X \in \mathbb{R}^d$ denote a feature vector, $Y \in \{0,1\}$ a binary label, and ${S \in \{s_1, \dots, s_n\}}$ a sensitive attribute, taken from some joint distribution $\mathbb{P}_{X,Y,S}$. 
The most common scenario in fairness-critical applications is \textit{binary} sensitive attribute ${S \in \{0, 1\}}$.
Traditional classification algorithms fit a classifier $f_\theta: \mathbb{R}^d \to \{0,1\}$ to predict the task label $Y$ from $X$.
The sensitive attribute $S$ is often statistically correlated with both the feature vector $X$ and the target label $Y$, raising fairness concerns.
A prevalent approach in state-of-the-art fairness-aware methods involves constructing fair representations $Z$ from $X$ prior to predicting $Y$.

\paragraph{Statistical Distance and Adversarial Evaluation}\label{sec:statistical_distance}

Since strict fairness criteria often entail accuracy trade-offs due to correlation between the sensitive $S$ and the task label $Y$, fair representations typically result in either performance reductions or leakage of the sensitive information in the classification.
Consequently, we adopt the concept of \textit{$\epsilon$-fairness}, where $\epsilon$ represents the statistical distance between the learned representations $\measure{Z \mid S}$.
This enables the application of bounds from Madras~et~al.~\cite{madras2018learning} relating statistical distances to several fairness metrics (see \cref{appendix:fairness_metrics}). 
Specifically, they consider the Total Variation (TV) distance, defined for distributions $\measure{0}$ and $\measure{1}$ as:
\begin{equation}
    \Delta(\measure{0}, \measure{1}) = \frac{1}{2}\int_{\mathcal{X}} |\measure{0}(x)-\measure{1}(x)|\,dx
    .
\end{equation}

Balunović et al.~\cite{balunovic2022fair} estimate the statistical distance directly. However, this can be extremely hard in the general case, and inaccuracy in the estimation might lead to a poor estimation of fairness. 
Given that the proposed method does not always impose specific distributional assumptions on $Z$, we uniformly adopt \textit{adversarial evaluation} to quantify fairness throughout the whole paper. 
Indeed, for an optimal adversarial classifier $f$, the statistical distance between conditional distributions can be precisely expressed as:
\begin{equation}
\label{eq:adversarial_accuracy_bound}
    \sup_{\mathcal{G}}\max_{g \in \mathcal{G}} \measure{}(Y = g(X)) = \frac{1}{2}\left(1+\Delta(\measure{Z \mid s=0}, \measure{Z \mid s=1})\right)
    .
\end{equation}

\textit{Adversarial Evaluation} should not be confused with
\textit{Adversarial Learning}.
Adversarial Learning learns a latent fair representation $Z$ from $X$ optimizing \cref{eq:adversarial_accuracy_bound} directly via a min-max training, which can lead to training instability \cite{feng2019learning, moyer2018invariant, elazar2018adversarial}.
Adversarial frameworks are typically robust only to the class of functions used during training.
Adversarial Evaluation simply assesses the fairness of a fixed representation $Z$ by quantifying how accurately a classifier (often an \ac{MLP}) can predict the sensitive attribute $S$ from $Z$.

\paragraph{Logistic Regression}
\label{sec:logistic_regression}
\ac{LR} is a popular classification algorithm used to model the probability of a binary outcome $Y\in\{0, 1\}$ as a function of a set of predictors $X\in\R^d$. The model is defined by the logistic function $\logistic{\beta}: \R^d \to (0, 1)$:
\begin{equation}
    \logistic{\beta}(x) 
    = \frac{1}{1 + e^{ -\beta\component{0} -\sum_{i=1}^d \beta\component{i} x\component{i}}},     
\end{equation}where the coefficients $\beta \in \R^{d+1}$ are the model parameters, with $\beta\component{0}$ denoting the bias term.
The model output $\logistic{\beta}(x)$ is used to learn $\measure{}(Y=1 \mid X=x)$ via maximum likelihood estimation. To achieve this, the objective function to be minimized is given by the negative log-likelihood:
\begin{equation}
\label{eq:log_likelihood}
    \mathcal{L}^\text{LR}(\beta) 
    = - \expectation[X, Y]{Y \log( \logistic{\beta}(X) ) + (1 - Y) \log(1 - \logistic{\beta}(X))}
\end{equation}

A key advantage of \ac{LR} over other families of classifiers, such as MLP, is the convexity of its objective function. Since this loss function is convex, it has a global optimum that can be provably reached. 
This property not only guarantees convergence but also enables the use of second-order optimization techniques such as Newton-Raphson or its variant, Iteratively Reweighted Least Squares. 
These methods leverage the Hessian of the loss function to achieve quadratic convergence, significantly accelerating the optimization compared to first-order methods like gradient descent.

\section{Fair Representations matching Characteristic Function}
\label{sec:fair_CF}

\begin{wrapfigure}{R}{0.5\textwidth}
  \vspace{-0.8cm}
  \begin{minipage}{0.51\textwidth}
  \begin{algorithm}[H]
    \caption{FmCF loss term}
    \label{alg:fmcf_short}
    \begin{algorithmic}[1]              
        \STATE \textbf{Input:} encoder $h_{\theta}$, predictor $f_\theta$, batch $\mathcal{B}=\{(x_i,y_i,s_i)\sim \measure{X,Y,S}\} $.
        \STATE $z \leftarrow h_{\theta}(x)$
        \STATE $\hat{y} \leftarrow f_{\theta}(z)$
        
        \FORALL{$s \in S$}
            \FORALL{$j \in \{1, \dots, k\}$}
                \STATE Sample $t_j\sim \measure{T}$
                \STATE $\charfun{\mathcal{N}}(t_j) \leftarrow e^{-{0.5\norm{t_j}^2}}$
                \STATE $\empcharfun{Z \mid s}(t_j) = \frac{1}{n}\sum_{i=1}^n e^{i \langle t_j, z_i\rangle}$
            \ENDFOR
        \ENDFOR
        \STATE $\mathcal{L}_{\text{CF}} \leftarrow \displaystyle\sum_{s\in \mathcal{S}}\frac{1}{k} \sum_{j=1}^k \absval{\charfun{\mathcal{N}}(t_j) -\empcharfun{Z \mid s}(t_j)}^2$
        \STATE $\mathcal{L} \leftarrow \mathcal{L}_{C}(\hat{y},y) + \alpha\,\mathcal{L}_{\text{CF}}$
    \end{algorithmic}
  \end{algorithm}
  \vspace{-0.3cm}
  \end{minipage}
\end{wrapfigure}

The concept of Characteristic Function has been extensively studied for a long time in the statistical testing literature \cite{lukacs1970characteristic}, but interest in its application to ML has only recently grown \cite{ansari2020characteristic}.
As discussed in \cref{sec:statistical_distance}, to incentivize fair representation, we need to match the different conditional distributions $\measure{Z \mid S}$, thus minimizing their relative statistical distance.
To achieve this, we propose the addition of a differentiable penalty term based on the Characteristic Function Distance. 
We refer to the overall approach as Fairness matching Characteristic Function (FmCF).

\paragraph{Characteristic Function Distance}
\label{sec:characteristic_function}

Given a random variable $X\in \R^d$, an alternative way to describe the distribution of $X$ is the \ac{CF} $\charfun{X}:\R^n\to \C$. 
Denoting by $\measure{X}$ the probability measure of $X$, the characteristic function $\charfun{X}$ is defined as:
\begin{equation}
\label{eq:characteristic_function_definition}
    \charfun{X}(t) 
    = \expectation[X]{e^{i \langle X, t \rangle}}
    = \int_{x\in\R^d} e^{i \langle x, t \rangle}  d\,\measure{X}  
    .
\end{equation}

The \ac{CF} is closely related to the notion of the Fourier Transform, and inherits several important properties. 
It always exists, it's bounded $|\varphi_X(t)| \le 1$, but most importantly, there is a one-to-one correspondence between probability measures and \ac{CF}s \cite{lukacs1970characteristic}. 
That is, given two random variables $X$ and $Y$, with probability measures $\measure{X}$ and $\measure{Y}$ then it holds that:

\begin{equation}
\label{eq:equal_phi_equal_measure}
\charfun{X} = \charfun{Y} \iff \measure{X} = \measure{Y}
.
\end{equation}

The \ac{CF} can be used to define a distance metric between probability measures \cite{ansari2020characteristic}. Given two random variables $X\in \R^d$, $Y\in \R^d$ with probability measures $\measure{X}$ and $\measure{Y}$, the \ac{CFD} is defined as:
\begin{equation}
\label{eq:characteristic_function_distance}
    \operatorname{CFD}_{\measure{T}}^2 \left( \measure{X}, \measure{Y} \right)
    =  \expectation[T]{\absval{\charfun{X}(T) -\charfun{Y}(T)}^2}
    .
\end{equation}
The value of $\operatorname{CFD}_{\measure{T}}$ depends on the weighting kernel $\measure{T}$. When $\operatorname{supp}(\measure{T})=\R^d$, then the \ac{CFD} is strict, in the sense that ${\operatorname{CFD}_{\measure{T}} \left( \measure{X}, \measure{Y} \right) = 0 \iff \measure{X} = \measure{Y}}$.
Common choices for $\measure{T}$ that ensure this property are the Normal and Laplace distributions.

\paragraph{Derivation of Penalty Term}
Instead of matching $\measure{Z \mid S}$ to each other, each $\measure{Z \mid S}$ is matched independently to a common target distribution. This simplifies and stabilizes the training procedure.

In this work, we use a standard Normal distribution as the target, although in principle, any distribution could be used.
The CF of the standard Normal distribution $\charfun{\mathcal{N}}$ is given by:
\begin{equation}
    \charfun{\mathcal{N}}(t)
    = \int_{x\in\R^d} (2\pi)^{\frac{-d}{2}} e^{-\frac{\norm{x}^2}{2}} e^{i \langle x, t \rangle} dx
    = e^{-\frac{\norm{t}^2}{2}}
    .
\end{equation}

\begin{figure}[t]
    \centering
    \includegraphics[width=0.99\textwidth]{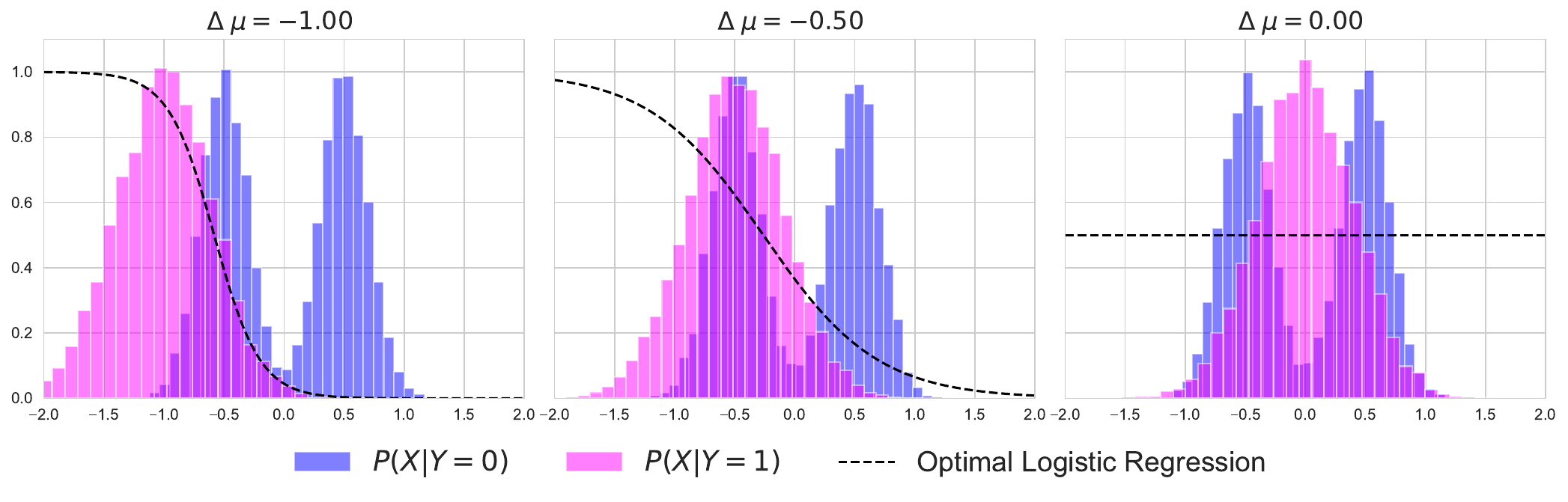}
    \caption{The closer $\mathbb{E}[X | y = 0]$ is to $\mathbb{E}[X | y = 1]$, the less predictive power the feature has, causing the \ac{LR} coefficient $\beta$ to approach 0.}
    \vskip -0.1cm
    \label{fig:example_LR}
\end{figure}

The \ac{CFD} between $\measure{Z \mid S}$ and the target distribution can be estimated using Monte Carlo sampling.
Given a batch of i.i.d. samples $z_1, \dots, z_n$ taken from $\measure{Z \mid S}$ and $t_1, \dots, t_j$ taken from the weighting kernel $\measure{T}$ the \ac{CFD} can be approximated as:
\begin{equation}\label{eq:empirical_characteristic_function_distance_definition}
    \operatorname{CFD}_{\measure{T}}^2\left( \measure{Z \mid S}, \measure{\mathcal{N}} \right)
    \approx \frac{1}{k} \sum_{j=1}^k \absval{\charfun{\mathcal{N}}(t_j) -\empcharfun{Z \mid S}(t_j)}^2
    .
\end{equation}
where $\empcharfun{Z \mid S}(t) = \frac{1}{n}\sum_{i=1}^n e^{i \langle t, z_i\rangle}$ denotes the empirical CF.
The total penalty term is obtained by summing the \ac{CFD} estimate across all different sensitive groups $\mathcal{S}$
\begin{equation}
    \mathcal{L}^{\text{CF}}
    = \sum_{s \in \mathcal{S}}\frac{1}{k} \sum_{j=1}^k \absval{\charfun{\mathcal{N}}(t_j) -\empcharfun{Z \mid S=s}(t_j)}^2.
\end{equation}

\paragraph{Comparison to Adversarial Learning and FNF} 
Similarly to FmCF, Adversarial Learning offers highly specialized representations; however, they have been shown to provide only an illusory sense of fairness \cite{moyer2018invariant, xu2020theory, song2019overlearning}, primarily due to the inherent characteristics of the underlying optimization problem. In contrast, the proposed approach seeks to learn representations that are strongly predictive of $Y$ while minimizing information about $S$, without the need for adversarial losses.

Compared to FNF \cite{balunovic2022fair}, FmCF offers multiple advantages compared to Normalizing Flows (NF).
Firstly, NFs require carefully designed architectures. 
To efficiently compute the Jacobian determinant, only specific layer structures are suitable. 
Additionally, to ensure invertibility, the dimensionality of the input vector $X$ must be preserved throughout the transformation. 
This constraint makes NFs computationally expensive for high-dimensional data.
In contrast, for FmCF, any function approximator can serve as an encoder. 
This architectural flexibility enables the learned representation $Z$ to focus solely on the information pertinent to the downstream task (e.g., classification).
Moreover, FmCF eliminates the need for sensitive information during deployment. 
Since the classifier internally debiases $X$, it does not require access to the sensitive attribute $S$ at test time. 
On the other hand, FNF not only necessitates access to the sensitive attribute but also requires training a separate Normalizing Flow for each sensitive class, significantly increasing training costs. 
This distinction is particularly important for minimizing potential discrimination and addressing concerns related to the explicit collection of sensitive data.
\newpage
\section{Fair Classification matching Sufficient Statistics}
\label{sec:fair_SS}
\begin{wrapfigure}{R}{0.5\textwidth}
  \vspace{-0.8cm}
  \begin{minipage}{0.51\textwidth}
  \begin{algorithm}[H]
    \caption{FmSS Training}
    \label{alg:fmss_short}
    \begin{algorithmic}[1]               
      \STATE \textbf{Input:} encoder $h_{\theta}$, linear regression $f_\beta$, batch $\mathcal{B}=\{(x_i,y_i,s_i)\sim \measure{X,Y,S}\} $.
        \STATE $z \leftarrow h_{\theta}(x)$
        \STATE $\hat{y} \leftarrow f_{\beta}(z)$
        \FORALL{$s \in S$}
          \STATE $\sigma_s^{2} \leftarrow \operatorname{Var}[Z \mid S=s]$
          \STATE $\mu_s \leftarrow \mathbb{E}[Z \mid S=s]$
        \ENDFOR
        \STATE $\mathcal{L}_{\text{KL}} \leftarrow
               \displaystyle\sum_{s\in \mathcal{S}} ||
               \sigma_s^{2} + \mu_s^{2} - 1 - \log\sigma_s^{2}||_1$
        \STATE $\mathcal{L} \leftarrow 
               \mathcal{L}_{C}(\hat{y},y) + \alpha\,\mathcal{L}_{\text{KL}}$
    \end{algorithmic}
  \end{algorithm}
  \vspace{-1.2cm}
  \end{minipage}
\end{wrapfigure}

The method introduced in \cref{sec:fair_CF} provides a versatile framework for learning specialized representations $Z$ from $X$ independent from $S$ for a general task. However, in practice, most fairness-aware scenarios involve classification \cite{wu2018fairness, madhavan2020fairness, zafar2017fairness}. By simplifying the approach, tailoring it to fair classification, we can provide provable post-hoc fairness guarantees. Furthermore, we can also make it computationally cheaper, relaxing the need for a Monte Carlo estimate of the CF.

\paragraph{Connection to Distribution Moments} The penalty term introduced in \cref{sec:fair_CF} involves sampling multiple points $t\sim\measure{T}$ from the weighting kernel where the empirical CF is evaluated at. The number of points needed for a reliable estimate of the \ac{CFD} grows rapidly in the number of dimensions \cite{ansari2020characteristic}.

In this section, we propose an alternative to reduce the number of points needed for the evaluation. Consider the multidimensional Maclaurin expansion of the CF:
\begin{equation}
    \charfun{X}(t)
    \approx \sum_{n=0}^\infty \sum_{\substack{n_1,\dots,n_d \\ \text{s.t. } \sum_i n_i = n}}^\infty
\frac{t\component{1}^{n_1} \cdots t\component{d}^{n_d}}
     {n_1!\cdots n_d!}
\left(
\frac{\partial^{n}\charfun{X}}
     {\partial t\component{1}^{n_1}\cdots\partial t\component{d}^{n_d}}
\right) \Big |_{t=0}.
\end{equation}
Assuming that $\charfun{X}$ is analytic, matching the derivatives evaluated at $t=0$ would be sufficient to ensure fairness, without the need for sampling from $\measure{T}$.
There exist statistical tools to estimate the full series (\cref{appendix:matching_all_moments}); however, they might lead to strong instabilities in training.

A pragmatic approach is to truncate the expansion to some order $N$. This objective does not guarantee that the two distributions are exactly equal; however, we will show that truncating to $N=2$ is sufficient to ensure fair classification.

In fact, there is a direct connection between the derivatives of the CF and the moments of a distribution.
\begin{theorem}
\label{thm:moments_from_phi}
    The $n$-derivative of the CF evaluated at $t=0$ is related to the $n$-th moment of the distribution:
    \begin{equation}
    \label{eq:generating_moments_with_phi}
        \frac{\partial^n  \charfun{X}}{\partial t\component{k_1}, \dots , \partial t\component{k_n}} \Big |_{t=0}
        = {i^n} \; \expectation[X]{X\component{k_1} \dots X\component{k_n}} 
        .
    \end{equation}
\end{theorem}

Similarly, the $n$-th empirical moment is equivalent to the $n$-th derivative of the empirical CF.

\paragraph{Simplified Penalty for Classification}
A generic \ac{MLP} with $M$ layers used for classification can be interpreted as $M-1$ layers of encoder $z = h_\theta(x)$, and a final layer of a \ac{LR} $\hat{y} = \logistic{\beta}(z)$. 
Therefore, by applying a fairness penalty to $z$, we can restrict the analysis to the family of \ac{LR} classifiers.

There exists an optimal condition under which \ac{LR} is provably fair. In particular, if the first moment $\expectation{Z \mid S}$ is the same for all $s\in S$, then such a representation $Z$ has no predictive power for \ac{LR} (i.e. $\beta=0$) in predicting $S$.

\begin{theorem}
\label{thm:optimal_LR}
    For a representation $Z \in \R^d$ and a binary sensitive attribute $S\in \{0, 1\}$, the optimal \ac{LR} classifier $\logistic{\beta^*}$ with $S$ as target is invariant to $Z\component{i}$ when
    $
    \expectation[Z\mid S]{Z\component{i}} = \expectation[Z]{Z\component{i}}
    .
    $
\end{theorem}

\Cref{thm:optimal_LR} shows that, building an intermediate representation $Z$ with $\expectation[Z \mid S]{Z} = \expectation[Z]{Z}$ is sufficient to ensure that \ac{LR} models do not use information about the sensitive attribute $S$. 

However, it is possible to derive an even stronger guarantee, analyzing what happens for ${\expectation[Z]{Z\component{i} \mid S} \approx \expectation[Z]{Z\component{i}}}$. 
In particular, \cref{thm:optimal_LR_convergence} highlights that the mean and variance of $\measure{Z \mid S}$ are enough to fully characterize the predictive power of a feature.
\begin{theorem}
    \label{thm:optimal_LR_convergence}
    The predictive power of a feature $Z\component{i}$ goes to zero the more its conditional expectation is close to the marginal one. That is, given 
    $\Omega =\frac{\norm{\expectation[Z \mid S]{Z\component{i}} - \expectation[Z]{Z\component{i}}}^2}{\expectation[Z]{Z\component{i}^2}}$ 
    \begin{equation}
        \lim_{\Omega \rightarrow 0} \beta^*\component{i} = 0
    \end{equation}
\end{theorem}

One empirical example of the consequences of \cref{thm:optimal_LR_convergence} is shown in \cref{fig:example_LR} where a Gaussian and a mixture of Gaussians are used to model $\measure{Z \mid S}$, and $\expectation[Z \mid S]{Z}$ gradually converge to the same value, while the variance is preserved.

Similarly to \cref{sec:fair_CF} we adopt a Gaussian target distribution for $\measure{Z\mid S}$
This is a particularly compelling choice from the lens of the maximum entropy principle\footnote{When constrained by fixed first and second moments, the Gaussian distribution uniquely maximizes entropy.}.

The fairness penalty term can be obtained from the Kullback–Leibler (KL) divergence between a standard Gaussian $\mathcal{N}(0, 1)$ and $\mathcal{N}(\mu, \sigma^2)$, which is given by 
$$
\operatorname{KL}(\mu, \sigma) = \frac{1}{2} \left( \sigma^2 + \mu^2 - 1 - \log \sigma^2 \right)
.
$$

Accordingly, the total penalty term can be obtained as
\begin{equation}
    \mathcal{L}^{\text{SS}}
    = \sum_{s\in \mathcal{S}} \norm{
        \operatorname{Var}[Z \mid S=s]
        +\expectation{Z \mid S=s} - 1 - \log \left(\operatorname{Var}[Z \mid S=s]\right)
    }
    .
\end{equation}
where $\operatorname{Var}[Z \mid S]$ and $\expectation{Z \mid S}$ denote the empirical moments.

\paragraph{Fairness Guarantees}
This approach fundamentally departs from popular methods in the literature, which primarily focus on minimizing $\Delta(\measure{Z \mid S=0}, \measure{Z \mid S=1})$, by permitting the sensitive attribute $S$ to be encoded within the learned representation. 
Crucially, it ensures that $S$ is embedded in a manner that renders it provably inaccessible to the classifier being trained. 
Leveraging the convexity of \ac{LR}, one can precisely quantify the $\epsilon$-fairness of a given representation $z$, thus establishing that no logistic classifier $\logistic{\beta}(x)$ can extract more than an $\epsilon$ amount of information regarding the sensitive attribute $s$. This constitutes a substantially stronger guarantee than adversarial evaluation, which lacks provable tightness and robustness.

\section{Experimental Evaluation}

\begin{wrapfigure}{R}{0.5\textwidth}
  \vspace{-2.0cm}
  \begin{minipage}{0.5\textwidth}
  \begin{figure}[H]
        \includegraphics[width=\textwidth]{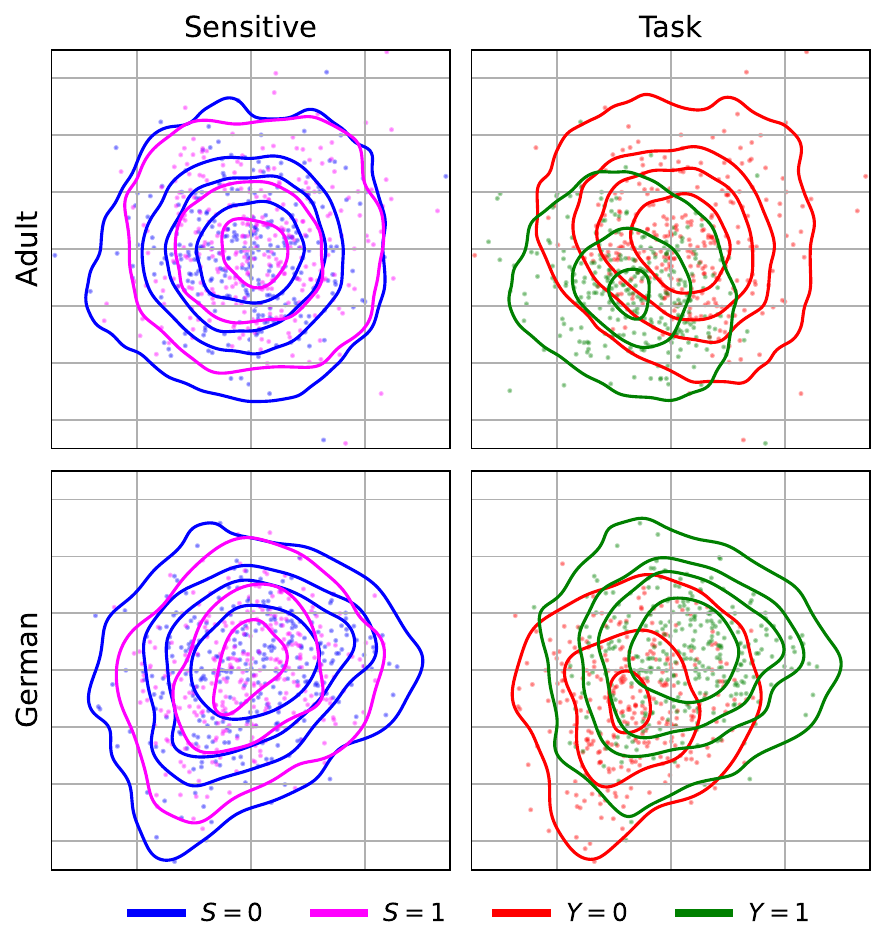}
        \caption{Latent distributions of $\measure{Z \mid S}$ and $\measure{Z|Y}$ obtained using FmCF.}
        \label{fig:learned_representations}
    \end{figure}
  \end{minipage}
  \vspace{-0.2cm}
\end{wrapfigure}

FNF \cite{balunovic2022fair} offers the strongest theoretical and empirical guarantees, making it the current state of the art in fair representation learning. Therefore, we primarily evaluate the performance of our proposed methods on the same suite of benchmarks, while also considering additional methods and datasets for a more comprehensive analysis. 
In all tested datasets, both proposed approaches match or surpass the performances of state-of-the-art approaches in terms of accuracy, while providing fairer representations. Surprisingly, even though the approach presented in \cref{sec:fair_SS} gives guarantees against the worst-case linear attacker, it still holds good performances against adversarial evaluations using deep neural networks. Other results on different setups can be found in \cref{appendix:other_results}, where we additionally report more Pareto plots using different hyperparameters for both FmSS (\cref{sec:fair_SS}) and FmCF (\cref{sec:fair_CF})

\begin{wraptable}{R}{0.5\textwidth}
  \vspace{0.0cm}
  \centering
  \caption{Comparison on Adult dataset.}
  \label{tab:adult}
  \begin{tabular}{ccccl}
    \toprule
    &&\multicolumn{2}{c}{Adv. Accuracy} \\
    \cmidrule(r){3-4}
    Model & Acc. & $g \in \mathcal{G}$ & $g \not\in \mathcal{G}$ \\
    \midrule
    Random & 75.20 & 50.00 & 50.00 \\
    AdvForgetting~\cite{jaiswal2020invariant} & 85.99 & 66.68 & 74.50 \\
    MaxEnt-ARL~\cite{roy2019mitigating} & 84.80 & 69.47 & 85.18 \\
    LAFTR~\cite{madras2018learning} & 86.09 & 72.05 & 84.58 \\
    FNF \cite{balunovic2022fair} & 84.43 & N/A & 59.56 \\
    \midrule
    FmCF (ours) & 85.01 & \textcolor{gray}{54.92} & 56.64 \\
    FmSS (ours) & 85.02 & 51.70 & \textcolor{gray}{57.47} \\
    \bottomrule
  \end{tabular}
  \vspace{0.cm}
\end{wraptable}
\paragraph{Adult Dataset} This dataset is the most fundamental benchmark in fairness, as almost all methods in literature are evaluated on it \cite{jaiswal2020invariant, roy2019mitigating, madras2018learning, balunovic2022fair}. It requires predicting the income of a person without considering their gender \cite{ding2021retiring}. For this dataset, we use the same setup as \cite{balunovic2022fair}, by evaluating our approach both on a classifier in the same family as the classifier used for the task, and one from a family of much more expressive ones, thus deeper and wider. 
This allows for a much tighter evaluation of the statistical distance $\Delta(\measure{Z \mid S=0}, \measure{Z \mid S=1})$. However, in \cref{appendix:other_results} we compare to other baselines that only offer guarantees on the trained family of functions. 
For the proposed methods in \cref{tab:adult} $g \in \mathcal{G}$ is the class of \ac{LR} classifiers, while $g \not\in \mathcal{G}$ are deep \ac{MLP}s (see \cref{appendix:training_details} for details). 
By construction, FmCF should be evaluated for $g \not \in \mathcal{G}$, while FmSS for $g \in \mathcal{G}$.

\begin{wraptable}{R}{0.4\textwidth}
  \vspace{-0.1cm}
  \centering
  \caption{Comparison on German dataset.}
    \label{tab:german}
    \begin{tabular}{ccc}
      \toprule
      Model & Acc. & Adv. Acc. \\
      \midrule
      Random & 69.00 & 70.00 \\
      VFAE~\cite{louizos2017variationalfairautoencoder} & 72.00 & 71.70 \\
      CIAFL~\cite{louizos2017variationalfairautoencoder} & 69.50 & 81.10 \\
      IRwAL~\cite{moyer2018invariant} & 71.00 & 69.80 \\
      \midrule
      FmCF (ours) & 74.60 & 69.00 \\
      FmSS (ours) & 74.20 & 69.00 \\
      \bottomrule
    \end{tabular}
  \vspace{-0.3cm}
\end{wraptable}

\paragraph{German dataset} 
The German dataset has been explored in the literature under different settings. In particular, \cite{moyer2018invariant} considers gender as the sensitive attribute, with the task of predicting income based on a set of personal features. To ensure a fair comparison, we evaluate our approach against several existing methods using the same experimental setup as in \cite{moyer2018invariant}, as presented in \cref{tab:german}. The proposed method outperforms the other approaches both in terms of accuracy and fairness, showcasing the effectiveness of the novel penalties. Further comparison under sensitive attributes is presented in \cref{appendix:other_results}.

\begin{figure}[t]
    \label{fig:pareto_plots}
    \includegraphics[width=0.97\linewidth]{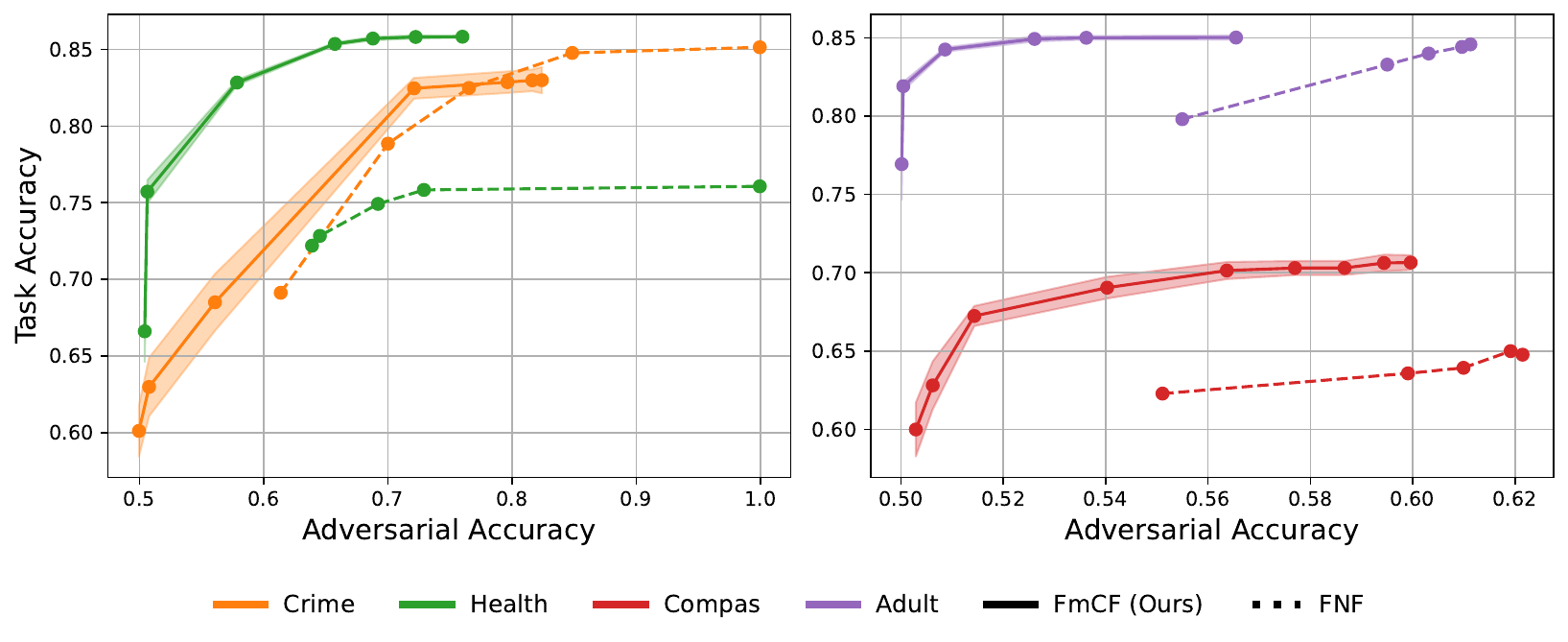}
    \hfill
    \caption{Pareto front between FmCF and FNF comparing task accuracy and fairness (95\% confidence intervals from 10 seeds).}
    \vspace{-0.5cm}
    \label{fig:side_by_side}
\end{figure}

\paragraph{Comparison with FNF \cite{balunovic2022fair}} Since \cite{balunovic2022fair} achieves the best balance between fairness and accuracy, in \cref{fig:pareto_plots} we compare our proposed method from \cref{sec:fair_CF} against all datasets used in their evaluation.
These include four distinct datasets: Crime and Health, which contain continuous features that NFs can handle directly, and Compas and Adult, which include discrete features requiring separate handling by Normalizing Flows. 

The plots demonstrate that our proposed method consistently outperforms or matches \cite{balunovic2022fair}, with the added advantage of not requiring the sensitive attribute during evaluation. Additionally, in \cref{appendix:other_results}, we report the Pareto fronts obtained by varying the latent dimension size, highlighting the impact on performance.

\paragraph{Visualization of Learned Representations} In order to have a better sense of what the CF approach presented in \cref{sec:fair_CF} is learning, we visualize the learned representation in \cref{fig:learned_representations}.\\
Indeed, it can be seen that for both datasets, the distribution $\measure{Z | S}$, which is the distribution of interest to evaluate the fairness of a representation, is very close to being a Gaussian. Thus, the CF is effectively enforcing the similarity to such a distribution. Instead, $\measure{Z|Y}$, which instead is the distribution of interest learn the downstream task, does not resemble a Gaussian distribution, and in particular, while $\measure{Z \mid S=0} \approx \measure{Z \mid S=1}$, $\measure{Z|Y=0} \not\approx \measure{Z|Y=1}$.

\section{Conclusions}
In this work, we introduce a novel framework for learning fair specialized representations using \ac{CFD}, addressing key limitations of existing adversarial and normalizing flow-based approaches. By leveraging CFs, our method ensures stable and efficient fairness by minimizing sensitive information leakage while maintaining high predictive performance. We also present a simplified version of the framework that allows for certifiable bounds on the amount of sensitive information used by downstream tasks. Experimental results show that our approaches consistently outperform or match state-of-the-art methods in terms of the fairness-accuracy trade-off across a range of benchmark datasets. Unlike many existing approaches, our method does not require access to sensitive attributes at inference time, making it more practical for real-world applications.
Our approach opens a new avenue for research in fair representation learning by offering a principled, non-adversarial alternative to existing methods. However, despite its simplicity and effectiveness, the proposed framework has certain limitations. Similar to many fairness-oriented algorithms, it encounters challenges in high-dimensional settings, where the \ac{CFD} tends to lose effectiveness. Moreover, the moment-based approach is inherently tailored to classification tasks, leaving opportunities for extending such approaches to regression contexts.
These limitations underscore the need for further research to improve the scalability and robustness of our method in such scenarios.
Additionally, current fairness benchmarks largely focus on low-dimensional datasets, limiting the ability to comprehensively evaluate and compare methods in more realistic and complex settings. Developing more sophisticated, high-dimensional fairness datasets would greatly enhance the evaluation of approaches like ours and drive further innovation in the field.

\section*{Broader Impact}
This research utilizes the Adult, COMPAS, German, Crime, and Health datasets, each of which is extensively acknowledged as a benchmark in the domain of machine learning fairness. Our study is conducted with stringent adherence to ethical standards and a commitment to transparency. By meticulously employing responsible methodologies, we strive to make substantive contributions to the advancement of AI ethics. This work underscores our dedication to fostering fairness and promoting socially responsible practices within the broader machine learning community.

\section*{Acknowledgements}
We thank the anonymous reviewers for their insightful comments and suggestions, which helped improve the clarity and quality of this work. We additionally thank Alessandro Fabris and Davide Dalle Pezze, for the valuable early-stage feedback and discussions on this paper.

\bibliographystyle{plain}
\bibliography{references}

\begin{thebibliography}{10}

\bibitem{ansari2020characteristic}
Abdul~Fatir Ansari, Jonathan Scarlett, and Harold Soh.
\newblock A characteristic function approach to deep implicit generative
  modeling.
\newblock In {\em Proceedings of the IEEE/CVF conference on computer vision and
  pattern recognition}, pages 7478--7487, 2020.

\bibitem{balunovic2022fair}
Mislav Balunovic, Anian Ruoss, and Martin Vechev.
\newblock Fair normalizing flows.
\newblock In {\em International Conference on Learning Representations}, 2022.

\bibitem{barocas-hardt-narayanan}
Solon Barocas, Moritz Hardt, and Arvind Narayanan.
\newblock {\em Fairness and Machine Learning: Limitations and Opportunities}.
\newblock MIT Press, 2023.

\bibitem{blake2024algorithmic}
Harrison Blake.
\newblock Algorithmic fairness: Developing methods to detect and mitigate bias
  in ai systems.
\newblock 11 2024.

\bibitem{brennan2009compas}
Tim Brennan, William Dieterich, and Beate Ehret.
\newblock Evaluating the predictive validity of the compas risk and needs
  assessment system.
\newblock {\em Criminal Justice and Behavior}, 2009.

\bibitem{brzezinski2024properties}
Dariusz Brzezinski, Julia Stachowiak, Jerzy Stefanowski, Izabela Szczech,
  Robert Susmaga, Sofya Aksenyuk, Uladzimir Ivashka, and Oleksandr Yasinskyi.
\newblock Properties of fairness measures in the context of varying class
  imbalance and protected group ratios.
\newblock {\em ACM Transactions on Knowledge Discovery from Data}, 2024.

\bibitem{chouldechova2017fair}
Alexandra Chouldechova.
\newblock Fair prediction with disparate impact: A study of bias in recidivism
  prediction instruments.
\newblock {\em Big data}, 5(2):153--163, 2017.

\bibitem{chwialkowski2015fast}
Kacper~P Chwialkowski, Aaditya Ramdas, Dino Sejdinovic, and Arthur Gretton.
\newblock Fast two-sample testing with analytic representations of probability
  measures.
\newblock {\em Advances in Neural Information Processing Systems}, 28, 2015.

\bibitem{cooper2024arbitrariness}
A.~Feder Cooper, Katherine Lee, Madiha~Zahrah Choksi, Solon Barocas,
  Christopher~De Sa, James Grimmelmann, Jon~M. Kleinberg, Siddhartha Sen, and
  Baobao Zhang.
\newblock Arbitrariness and social prediction: The confounding role of variance
  in fair classification.
\newblock In Michael~J. Wooldridge, Jennifer~G. Dy, and Sriraam Natarajan,
  editors, {\em Thirty-Eighth {AAAI} Conference on Artificial Intelligence,
  {AAAI} 2024, Thirty-Sixth Conference on Innovative Applications of Artificial
  Intelligence, {IAAI} 2024, Fourteenth Symposium on Educational Advances in
  Artificial Intelligence, {EAAI} 2014, February 20-27, 2024, Vancouver,
  Canada}, pages 22004--22012. {AAAI} Press, 2024.

\bibitem{cornacchia2023auditing}
Giandomenico Cornacchia, Vito~Walter Anelli, Giovanni~Maria Biancofiore,
  Fedelucio Narducci, Claudio Pomo, Azzurra Ragone, and Eugenio~Di Sciascio.
\newblock Auditing fairness under unawareness through counterfactual reasoning.
\newblock {\em Inf. Process. Manag.}, 60(2):103224, 2023.

\bibitem{creager2019flexibly}
Elliot Creager, David Madras, J{\"o}rn-Henrik Jacobsen, Marissa Weis, Kevin
  Swersky, Toniann Pitassi, and Richard Zemel.
\newblock Flexibly fair representation learning by disentanglement.
\newblock In {\em International conference on machine learning}, pages
  1436--1445. PMLR, 2019.

\bibitem{davrazos2025evaluting}
Gregory Davrazos and Sotiris Kotsiantis.
\newblock Evaluating fairness strategies in educational data mining: A
  comparative study of bias mitigation techniques.
\newblock {\em Electronics}, 14, 05 2025.

\bibitem{denis2021fairness}
Christophe Denis, Romuald Elie, Mohamed Hebiri, and Fran{\c{c}}ois Hu.
\newblock Fairness guarantee in multi-class classification.
\newblock {\em arXiv preprint arXiv:2109.13642}, 2021.

\bibitem{ding2021retiring}
Frances Ding, Moritz Hardt, John Miller, and Ludwig Schmidt.
\newblock Retiring adult: New datasets for fair machine learning.
\newblock {\em Advances in neural information processing systems},
  34:6478--6490, 2021.

\bibitem{dualgorithmic}
Mengnan Du, Lu~Cheng, and Dejing Dou.
\newblock Algorithmic fairness in machine learning.

\bibitem{du2020fairness}
Mengnan Du, Fan Yang, Na~Zou, and Xia Hu.
\newblock Fairness in deep learning: A computational perspective.
\newblock {\em IEEE Intelligent Systems}, 36(4):25--34, 2020.

\bibitem{dua2019uci}
Dheeru Dua and Casey Graff.
\newblock {UCI} machine learning repository.
\newblock \url{https://archive.ics.uci.edu/ml}, 2017.
\newblock Accessed: 2025-05-12.

\bibitem{dwork2012fairness}
Cynthia Dwork, Moritz Hardt, Toniann Pitassi, Omer Reingold, and Richard Zemel.
\newblock Fairness through awareness.
\newblock In {\em Proceedings of the 3rd innovations in theoretical computer
  science conference}, pages 214--226, 2012.

\bibitem{edwards2015censoring}
Harrison Edwards and Amos Storkey.
\newblock Censoring representations with an adversary.
\newblock {\em arXiv preprint arXiv:1511.05897}, 2015.

\bibitem{elazar2018adversarial}
Yanai Elazar and Yoav Goldberg.
\newblock Adversarial removal of demographic attributes from text data.
\newblock {\em arXiv preprint arXiv:1808.06640}, 2018.

\bibitem{fabris2024fairness}
Alessandro Fabris, Nina Baranowska, Matthew~J Dennis, David Graus, Philipp
  Hacker, Jorge Saldivar, Frederik Zuiderveen~Borgesius, and Asia~J Biega.
\newblock Fairness and bias in algorithmic hiring.
\newblock {\em ACM Transactions on Intelligent Systems and Technology}, 2024.

\bibitem{feng2019learning}
Rui Feng, Yang Yang, Yuehan Lyu, Chenhao Tan, Yizhou Sun, and Chunping Wang.
\newblock Learning fair representations via an adversarial framework.
\newblock {\em arXiv preprint arXiv:1904.13341}, 2019.

\bibitem{gajane2018formalizing}
Pratik Gajane and Mykola Pechenizkiy.
\newblock On formalizing fairness in prediction with machine learning, 2018.

\bibitem{gillis2024operationalizing}
Talia~B. Gillis, Vitaly Meursault, and Berk Ustun.
\newblock Operationalizing the search for less discriminatory alternatives in
  fair lending.
\newblock In {\em The 2024 {ACM} Conference on Fairness, Accountability, and
  Transparency, FAccT 2024, Rio de Janeiro, Brazil, June 3-6, 2024}, pages
  377--387. {ACM}, 2024.

\bibitem{golz2019paradoxes}
Paul G{\"o}lz, Anson Kahng, and Ariel~D Procaccia.
\newblock Paradoxes in fair machine learning.
\newblock {\em Advances in Neural Information Processing Systems}, 32, 2019.

\bibitem{gupta2021controllable}
Umang Gupta, Aaron~M Ferber, Bistra Dilkina, and Greg Ver~Steeg.
\newblock Controllable guarantees for fair outcomes via contrastive information
  estimation.
\newblock In {\em Proceedings of the AAAI Conference on Artificial
  Intelligence}, volume~35, pages 7610--7619, 2021.

\bibitem{hardt2016equality}
Moritz Hardt, Eric Price, and Nati Srebro.
\newblock Equality of opportunity in supervised learning.
\newblock {\em Advances in neural information processing systems}, 29, 2016.

\bibitem{jaiswal2020invariant}
Ayush Jaiswal, Daniel Moyer, Greg Ver~Steeg, Wael AbdAlmageed, and Premkumar
  Natarajan.
\newblock Invariant representations through adversarial forgetting.
\newblock In {\em Proceedings of the AAAI Conference on Artificial
  Intelligence}, volume~34, pages 4272--4279, 2020.

\bibitem{jiang2020wasserstein}
Ray Jiang, Aldo Pacchiano, Tom Stepleton, Heinrich Jiang, and Silvia Chiappa.
\newblock Wasserstein fair classification.
\newblock In {\em Uncertainty in artificial intelligence}, pages 862--872.
  PMLR, 2020.

\bibitem{kingma2014adam}
Diederik~P Kingma and Jimmy Ba.
\newblock Adam: A method for stochastic optimization.
\newblock {\em arXiv preprint arXiv:1412.6980}, 2014.

\bibitem{locatello2019fairness}
Francesco Locatello, Gabriele Abbati, Thomas Rainforth, Stefan Bauer, Bernhard
  Sch{\"o}lkopf, and Olivier Bachem.
\newblock On the fairness of disentangled representations.
\newblock {\em Advances in neural information processing systems}, 32, 2019.

\bibitem{louizos2017variationalfairautoencoder}
Christos Louizos, Kevin Swersky, Yujia Li, Max Welling, and Richard Zemel.
\newblock The variational fair autoencoder, 2017.

\bibitem{lukacs1970characteristic}
Eugene Lukacs.
\newblock Characteristic functions.
\newblock {\em (No Title)}, 1970.

\bibitem{madhavan2020fairness}
Ramanujam Madhavan and Mohit Wadhwa.
\newblock Fairness-aware learning with prejudice free representations.
\newblock In {\em Proceedings of the 29th ACM International Conference on
  Information \& Knowledge Management}, CIKM '20, page 2137–2140, New York,
  NY, USA, 2020. Association for Computing Machinery.

\bibitem{madras2018learning}
David Madras, Elliot Creager, Toniann Pitassi, and Richard Zemel.
\newblock Learning adversarially fair and transferable representations.
\newblock In {\em International Conference on Machine Learning}, pages
  3384--3393. PMLR, 2018.

\bibitem{mcnamara2019costs}
Daniel McNamara, Cheng~Soon Ong, and Robert~C. Williamson.
\newblock Costs and benefits of fair representation learning.
\newblock In {\em Proceedings of the 2019 AAAI/ACM Conference on AI, Ethics,
  and Society}, AIES '19, page 263–270, New York, NY, USA, 2019. Association
  for Computing Machinery.

\bibitem{moyer2018invariant}
Daniel Moyer, Shuyang Gao, Rob Brekelmans, Aram Galstyan, and Greg Ver~Steeg.
\newblock Invariant representations without adversarial training.
\newblock {\em Advances in neural information processing systems}, 31, 2018.

\bibitem{obermeyer2019dissecting}
Ziad Obermeyer, Brian Powers, Christine Vogeli, and Sendhil Mullainathan.
\newblock Dissecting racial bias in an algorithm used to manage the health of
  populations.
\newblock {\em Science}, 366(6464):447--453, Oct 2019.

\bibitem{pitoura2022fairness}
Evaggelia Pitoura, Kostas Stefanidis, and Georgia Koutrika.
\newblock Fairness in rankings and recommendations: an overview.
\newblock {\em The VLDB Journal}, pages 1--28, 2022.

\bibitem{roh2020fr}
Yuji Roh, Kangwook Lee, Steven Whang, and Changho Suh.
\newblock Fr-train: A mutual information-based approach to fair and robust
  training.
\newblock In {\em International Conference on Machine Learning}, pages
  8147--8157. PMLR, 2020.

\bibitem{roy2019mitigating}
Proteek~Chandan Roy and Vishnu~Naresh Boddeti.
\newblock Mitigating information leakage in image representations: A maximum
  entropy approach.
\newblock In {\em Proceedings of the IEEE/CVF Conference on Computer Vision and
  Pattern Recognition}, pages 2586--2594, 2019.

\bibitem{R_z_2021}
Tim Räz.
\newblock Group fairness: Independence revisited.
\newblock In {\em Proceedings of the 2021 ACM Conference on Fairness,
  Accountability, and Transparency}, FAccT ’21, page 129–137. ACM, March
  2021.

\bibitem{seyyedkalantari2020chexclusionfairnessgapsdeep}
Laleh Seyyed-Kalantari, Guanxiong Liu, Matthew McDermott, Irene~Y. Chen, and
  Marzyeh Ghassemi.
\newblock Chexclusion: Fairness gaps in deep chest x-ray classifiers, 2020.

\bibitem{siddique2023survey}
Sunzida Siddique, Mohd~Ariful Haque, Roy George, Kishor~Datta Gupta, Debashis
  Gupta, and Md~Jobair~Hossain Faruk.
\newblock Survey on machine learning biases and mitigation techniques.
\newblock {\em Digital}, 4(1):1--68, 2023.

\bibitem{song2019overlearning}
Congzheng Song and Vitaly Shmatikov.
\newblock Overlearning reveals sensitive attributes.
\newblock {\em arXiv preprint arXiv:1905.11742}, 2019.

\bibitem{song2019learning}
Jiaming Song, Pratyusha Kalluri, Aditya Grover, Shengjia Zhao, and Stefano
  Ermon.
\newblock Learning controllable fair representations.
\newblock In {\em The 22nd International Conference on Artificial Intelligence
  and Statistics}, pages 2164--2173. PMLR, 2019.

\bibitem{tucker2022prototype}
Mycal Tucker and Julie~A Shah.
\newblock Prototype based classification from hierarchy to fairness.
\newblock In {\em International Conference on Machine Learning}, pages
  21884--21900. PMLR, 2022.

\bibitem{vetro2021data}
Antonio Vetr{\`{o}}, Marco Torchiano, and Mariachiara Mecati.
\newblock A data quality approach to the identification of discrimination risk
  in automated decision making systems.
\newblock {\em Gov. Inf. Q.}, 38(4):101619, 2021.

\bibitem{wang2022brief}
Xiaomeng Wang, Yishi Zhang, and Ruilin Zhu.
\newblock A brief review on algorithmic fairness.
\newblock {\em Management System Engineering}, 1(1):7, 2022.

\bibitem{wu2018fairness}
Yongkai Wu, Lu~Zhang, and Xintao Wu.
\newblock Fairness-aware classification: Criterion, convexity, and bounds.
\newblock {\em arXiv preprint arXiv:1809.04737}, 2018.

\bibitem{xie2017controllable}
Qizhe Xie, Zihang Dai, Yulun Du, Eduard Hovy, and Graham Neubig.
\newblock Controllable invariance through adversarial feature learning.
\newblock {\em Advances in neural information processing systems}, 30, 2017.

\bibitem{xu2020theory}
Yilun Xu, Shengjia Zhao, Jiaming Song, Russell Stewart, and Stefano Ermon.
\newblock A theory of usable information under computational constraints.
\newblock {\em arXiv preprint arXiv:2002.10689}, 2020.

\bibitem{Yu_2024}
Zhe Yu, Joymallya Chakraborty, and Tim Menzies.
\newblock Fairbalance: How to achieve equalized odds with data pre-processing.
\newblock {\em IEEE Transactions on Software Engineering}, 50(9):2294–2312,
  September 2024.

\bibitem{zafar2017fairness}
Muhammad~Bilal Zafar, Isabel Valera, Manuel~Gomez Rogriguez, and Krishna~P.
  Gummadi.
\newblock {Fairness Constraints: Mechanisms for Fair Classification}.
\newblock In Aarti Singh and Jerry Zhu, editors, {\em Proceedings of the 20th
  International Conference on Artificial Intelligence and Statistics},
  volume~54 of {\em Proceedings of Machine Learning Research}, pages 962--970.
  PMLR, 20--22 Apr 2017.

\bibitem{zemel2013learning}
Rich Zemel, Yu~Wu, Kevin Swersky, Toni Pitassi, and Cynthia Dwork.
\newblock Learning fair representations.
\newblock In {\em International conference on machine learning}, pages
  325--333. PMLR, 2013.

\bibitem{zhong2024intrinsic}
Meiyu Zhong and Ravi Tandon.
\newblock Intrinsic fairness-accuracy tradeoffs under equalized odds.
\newblock In {\em 2024 IEEE International Symposium on Information Theory
  (ISIT)}, pages 220--225. IEEE, 2024.

\end{thebibliography}

\newpage
\appendix
\section{Appendix}
This appendix provides additional theoretical derivations, algorithmic details, and extended empirical results to support the main paper. It is organized as follows:

\begin{itemize}
  \item \textbf{\Cref{appendix:proofs}}: Complete derivations for Theorems \ref{thm:moments_from_phi}, \ref{thm:optimal_LR}, and \ref{thm:optimal_LR_convergence}.
  
  \item \textbf{\Cref{appendix:mmd}}: Analysis of MMD, its relation to characteristic‐function distances, and links to adversarial representation learning.
  
  \item \textbf{\Cref{appendix:matching_all_moments}}: Discussion of full moment‐matching, the Russian‐Roulette estimator for unbiased series truncation, and practical considerations.
  
  \item \textbf{\Cref{appendix:fairness_metrics}}: Formal definitions and commentary on Demographic Parity, Equalized Odds, and other group‐fairness criteria.
  
  \item \textbf{\Cref{appendix:other_results}}: Extended comparisons against baselines, Pareto‐front ablations over latent dimensions, and robustness checks.
  
  \item \textbf{\Cref{appendix:datasets}}: Detailed descriptions of all benchmark datasets, preprocessing pipelines, and protected attribute encoding.
  
  \item \textbf{\Cref{appendix:training_details}}: Implementation specifics, optimizer settings, hyperparameters, and hardware configuration.
\end{itemize}

\subsection{Proofs of Theorems}\label{appendix:proofs}
\begin{proof}{of \cref{thm:moments_from_phi}}

Taking the derivative inside the expectation
\begin{align}
    \frac{\partial^n  \charfun{X}}{\partial t\component{k_1}, \dots , \partial t\component{k_n}}
    &= \frac{\partial^n  \expectation[X]{e^{i \langle X, t \rangle}}}{\partial t\component{k_1}, \dots , \partial t\component{k_n}}  \\
    &= \expectation[X]{\frac{\partial^n  e^{i \langle X, t \rangle}}{\partial t\component{k_1}, \dots , \partial t\component{k_n}}} \\
    &= \expectation[X]{i^n  X\component{k_1} \dots X\component{k_n} e^{i \langle X, t \rangle}} 
    .
\end{align}
Evaluating both sides at $t=0$ concludes the proof:
\end{proof}

\begin{proof}{of \cref{thm:optimal_LR}}

    Defining $x\component{0}=1$, the derivative of the logistic function can be expressed as
    \begin{align}  
        \frac{\partial \logistic{\beta}(x)}{\partial\beta\component{i}}
        = x\component{i} \, \logistic{\beta}(x) (1-\logistic{\beta}(x)) 
        .
    \end{align}
    
    The derivative of the log-likelihood function in \cref{eq:log_likelihood} can be expressed as
    \begin{align}  
        \frac{\partial\mathcal{L}^\text{LR}(\beta)}{\partial \beta\component{i}} 
        &= - \frac{\partial}{\partial \beta\component{i}} \expectation[X, Y]{Y \log( \logistic{\beta}(X) ) + (1 - Y) \log(1 - \logistic{\beta}(X))} \\
        &= - \expectation[X, Y]{
            Y \frac{1}{\logistic{\beta}(X)}\frac{\partial \logistic{\beta}(X)}{\partial\beta\component{i}}
            -(1-Y) \frac{1}{(1-\logistic{\beta}(X))}\frac{\partial \logistic{\beta}(X)}{\partial\beta\component{i}}
        } \\
        &= - \expectation[X, Y]{
            Y \, x\component{i} \, (1-\logistic{\beta}(x)) 
            - (1-Y) \, x\component{i} \, \logistic{\beta}(x) 
        } \\
        &= - \expectation[X, Y]{
            x\component{i} \left( Y - \logistic{\beta}(x) \right)
        }
        .
    \end{align}
    

    At the optimum $\beta^*$, the gradient of the log-likelihood is zero.
    Considering the partial derivative with respect to $\beta\component{0}$
    \begin{align}  
        0=\frac{\partial \mathcal{L}^\text{LR} (\beta^*)}{\partial \beta\component{0}}
        &= -\expectation[X,Y]{Y - \logistic{\beta^*}(X)} \\
        &= - \expectation[Y]{Y} + \expectation[X]{\logistic{\beta^*}(X)}
    \end{align}
    showing that $\expectation[Y]{Y} = \expectation[X]{\logistic{\beta^*}(X)}$.
    
    Considering the partial derivative with respect to $\beta\component{i}$ for $i \neq 0$
    \begin{align}  
        0=\frac{\partial \mathcal{L}^\text{LR} (\beta^*)}{\partial\beta\component{i}} 
        &= -\expectation[X,Y]{X\component{i} \left(Y - \logistic{\beta^*}(X)\right)} \\
        &= \expectation[X]{X\component{i} \; \logistic{\beta^*}(X)} - \expectation[Y]{Y \; \expectation[X\mid Y]{X\component{i}}}
        .
    \end{align}

    Substituting $\expectation[X]{\logistic{\beta^*}(X)} = \expectation[Y]{Y}$ and the hypotesis $\expectation[X \mid Y]{X\component{i}}] = \expectation[X]{X\component{i}}$ 
    \begin{align}     
        0=\frac{\partial \mathcal{L}^\text{LR} (\beta^*)}{\partial\beta\component{i}} 
        &=\expectation[X]{X\component{i} \; \logistic{\beta^*}(X)} - \expectation[Y]{Y} \; \expectation[X]{X\component{i}} \\
        &=\expectation[X]{X\component{i} \; \logistic{\beta^*}(X)} - \expectation[X]{\logistic{\beta^*}(X)} \; \expectation[X]{X\component{i}} \\
        &= \operatorname{Cov}\left( X\component{i} ; \logistic{\beta^*}(X) \right)
    \end{align}  

    Unless $X\component{i}$ is constant, the condition ${\operatorname{Cov}\left( X\component{i} ; \logistic{\beta^*}(X) \right) = 0}$ is obtained for $\beta^*\component{i}=0$, making $\logistic{\beta^*}(X)$ independent of $X\component{i}$. 
\end{proof}

\begin{proof}{\cref{thm:optimal_LR_convergence}}

    Following the same steps of the proof for \cref{thm:optimal_LR}, we can show that $\expectation[X]{\logistic{\beta^*}(X)} = \expectation[Y]{Y}$ and ${\expectation[X]{X\component{i} \; \logistic{\beta^*}(X)} = \expectation[Y]{Y \; \expectation[X\mid Y]{X\component{i}}}}$.

    Therefore, the correlation $\operatorname{Corr}\left( X\component{i} ; \logistic{\beta^*}(X) \right)$ can be expressed as:
    \begin{align}  
        \operatorname{Corr}\left( X\component{i} ; \logistic{\beta^*}(X) \right)
        & = \frac{\expectation[X]{\left(X\component{i} -\expectation[X]{X\component{i}}\right) \left(\logistic{\beta^*}(X) -\expectation[X]{\logistic{\beta^*}(X)}\right)}
        }{\sqrt{\expectation[X]{X\component{i}^2} \expectation[X]{\logistic{\beta^*}^2(X)}}} \\
        & = \frac{\expectation[X]{X\component{i} \logistic{\beta^*}(X)} -\expectation[X]{X\component{i}}\expectation[X]{\logistic{\beta^*}(X)}
        }{\sqrt{\expectation[X]{X\component{i}^2} \expectation[X]{\logistic{\beta^*}^2(X)}}}
        .
    \end{align}

    Using the triangle inequality $\expectation[X]{\logistic{\beta}(X)^2} \ge \expectation[X]{\logistic{\beta}(X)}^2$ and substituting ${\logistic{\beta}(X) = \expectation[Y]{Y}}$ and ${\expectation[X]{X\component{i} \; \logistic{\beta^*}(X)} = \expectation[Y]{Y \; \expectation[X\mid Y]{X\component{i}}}}$:
    \begin{align}  
        \operatorname{Corr}\left( X\component{i} ; \logistic{\beta^*}(X) \right)
        & \leq \frac{\expectation[X]{X\component{i} \logistic{\beta^*}(X)} -\expectation[X]{X\component{i}} \expectation[X]{\logistic{\beta^*}(X)}
        }{\sqrt{\expectation[X]{X\component{i}^2}} \expectation[X]{\logistic{\beta^*}(X)}} \\
        & = \frac{\expectation[Y]{Y \; \expectation[X\mid Y]{X\component{i}}} -\expectation[X]{X\component{i}}\expectation[Y]{Y}
        }{\sqrt{\expectation[X]{X\component{i}^2}} \expectation[Y]{Y}} \\
        & = \frac{\expectation[Y]{Y \left(\expectation[X\mid Y]{X\component{i}} -\expectation[X]{X\component{i}}\right)}
        }{\sqrt{\expectation[X]{X\component{i}^2}} \expectation[Y]{Y}}
        .
    \end{align}

    Assuming $\Omega =\frac{\norm{\expectation[X\mid Y]{X\component{i}} -\expectation[X]{X\component{i}}}^2}{\expectation[X]{X\component{i}^2}} < \epsilon$ we can bound the correlation with
    \begin{align}  
        \operatorname{Corr}\left( X\component{i} ; \logistic{\beta^*}(X) \right) \leq \sqrt{\epsilon}
        .
    \end{align}
    Showing that $\lim_{\Omega \rightarrow 0} \operatorname{Corr}\left( X\component{i} ; \logistic{\beta^*}(X) \right) = 0$. When $X\component{i}$ is not constant, this also implies ${\lim_{\Omega \rightarrow 0} \beta^*\component{i} = 0}$
\end{proof}

\subsection{Connections to Maximum Mean Discrepancy}\label{appendix:mmd}
The Maximum Mean Discrepancy (MMD) is a popular distance measure between distributions. Given a Reproducing Kernel Hilber Space (RKHS) $\mathcal{H}$, it is defined as follows
\begin{equation}
    \operatorname{MMD}_\mathcal{H}\left(\measure{X}, \measure{Y}\right)
    = \sup_{\norm{f}_\mathcal{H} \leq 1} \expectation{f(X)} - \expectation{f(Y)}
    .
\end{equation}

\paragraph{Connections from Characteristic Function} 
For random variables $X\in \R^d$, $Y\in \R^d$, the MMD can be expressed in terms of the CFs \cite{chwialkowski2015fast}.
\begin{equation}
    \operatorname{MMD}_\mathcal{H}^2\left(\measure{X}, \measure{Y}\right)
    = \int_{\R^d} \absval{\charfun{X}(t) - \charfun{Y}(t)}^2 w(t) dt
    .
\end{equation}
where the weighting function $w(t)$ is the inverse Fourier Transform of the kernel of $\mathcal{H}$. 
When $w$ is a probability density function, then the integral can be interpreted as an expectation, and
\begin{equation}
    \operatorname{MMD}_\mathcal{H}\left(\measure{X}, \measure{Y}\right)
    =\operatorname{CFD}_{\measure{T}} \left( \measure{X}, \measure{Y} \right)
\end{equation}

\paragraph{Connections from Sufficient Statistics} 
The formulation presented in \cref{sec:fair_SS} shares some similarities with Maximum Mean Discrepancy (MMD) used by VFAE. Indeed, MMD states that given two distributions, they are equal as long as the maximum distance between their expected value under a transformation $f$ is zero.

Thus, if $\operatorname{MMD}_\mathcal{H}\left(\measure{X}, \measure{Y}\right)=0$ then $\measure{X} = \measure{Y}$. This is indeed exploited by Adversarial Learning, which tries to find $f$ by means of a min-max optimization. 
However, as shown by \cite{balunovic2022fair}, Adversarial Learning fails as finding such $f$ is extremely hard, due to the fact that the two distributions are constantly changing based on the current $f$. 
Therefore, if $\operatorname{MMD}_\mathcal{H}\left(\measure{X}, \measure{Y}\right) \ne 0$, no provable guarantee on the fairness can be given. Instead, if $f$ is in the family of \ac{LR}, \cref{thm:optimal_LR_convergence} shows that the second moment bounds the predictive power of such a representation.

This is a weaker condition than the one that \cite{balunovic2022fair}, Adversarial Learning, CVAE, uses to learn, by matching the distribution. Indeed, $\measure{Z \mid S=0} = \measure{Z \mid S=1} \Rightarrow \mathbb{E}[Z \mid S=1] = \mathbb{E}[Z \mid S=0]$ but not the other way around.

Yet, given \ac{LR} has a convex loss function, the optimal adversarial classifier can be provably trained to convergence with a second-order optimizer, thus giving guarantees on the amount of sensitive $S$ that can at most be used in the classification.



\subsection{Matching All Moments}\label{appendix:matching_all_moments}
The approaches presented in \cref{sec:fair_CF,sec:fair_SS} can be reconducted to matching the moment generating function, or directly the moments of a distribution.
Indeed, matching all moments of two distributions with bounded support ensures that they are equal almost everywhere. 
Even though it is trivial to force the support of $Z$ to be bounded, directly applying moment matching to the distributions $\measure{Z \mid S}$ might be impractical, as all moments are needed to characterize a distribution, which is computationally unfeasible.
Furthermore, truncating to the $k$-th moment leads to a biased estimation, and, in the context of this work, potential leakage of the sensitive attribute $S$.

One powerful tool for unbiased estimation of the complete infinite series of moments is the \textit{Russian Roulette estimator}. Consider an infinite series of the form $Y = \sum_{i=1}^\infty Y_i$.
The Russian Roulette estimator randomly chooses a truncation point $N$ and calculates the partial sum up to $N$, while adjusting the weight of the computed sum to ensure that the estimator remains unbiased:
\begin{equation}
    \hat{Y} = \sum_{i=1}^{N} \frac{Y_i}{P(N \geq i)}
\end{equation}
While the Russian Roulette estimator is unbiased, it can result in high variance if the truncation probabilities are not chosen carefully. Therefore, balancing computational efficiency and variance is essential in practical applications.

Instead, the proposed approaches presented in \cref{sec:fair_CF,sec:fair_SS} use a more principled and computationally stable way of matching two distributions.

\subsection{Additional Fairness Metrics}\label{appendix:fairness_metrics}

Although various fairness criteria have been proposed, balancing the fairness-accuracy trade-off remains challenging.

Demographic Parity (DP) is a widely used group fairness criterion that requires the outcome of a decision-making process to be independent of a sensitive attribute such as race, gender, or age \cite{dwork2012fairness, denis2021fairness}. Let $h_{\theta}$ denote a representation function that maps inputs $X$ to a latent space $Z = h_{\theta}(X)$, and let $f_{\theta}$ denote a classifier applied to this representation, producing predictions $\hat{Y} = f_{\theta}(h_{\theta}(X))$. Then, DP requires the probability of receiving a favorable outcome (e.g., $\hat{Y} = 1$) to be the same across all demographic groups $s \in \mathcal{S}$:
\begin{equation}
\label{eq:dp_ideal}
\mathbb{P}(\hat{Y} = 1 \mid S=s) = \mathbb{P}(\hat{Y} = 1) \quad \forall s \in \mathcal{S}
.
\end{equation}
This states that the positive outcome rate for any group $s$ should be equal to the overall positive outcome rate in the population. Since demographic parity is independent of the ground truth labels, it is especially salient in contexts where reliable ground truth information is hard to obtain and a positive outcome is desirable, including employment, credit, and criminal justice \cite{du2020fairness,gajane2018formalizing}.

To quantify deviations from this ideal, we measure the Demographic Parity Difference (DPD). In binary group settings (e.g., involving an advantaged group $a$ and a disadvantaged group $d$), it is defined as:
\begin{equation}
\label{eq:dpd_binary}
\operatorname{DPD} = \mathbb{P}(\hat{Y} = 1 \mid S=a) - \mathbb{P}(\hat{Y} = 1 \mid S=d)
.
\end{equation}

In real-world scenarios, sensitive attributes are often multi-class (e.g., race with more than two categories). In these cases, generalizations for measuring DPD violation include:
\begin{itemize}
    \item The \emph{Maximum Pairwise Difference}, capturing the largest disparity in positive outcome rates between any two groups:
    \begin{equation}
    \label{eq:dp_max_pair}
    \operatorname{DPD} = \max_{s_i, s_j \in \mathcal{S}} \left| \mathbb{P}(\hat{Y} = 1 \mid S = s_i) - \mathbb{P}(\hat{Y} = 1 \mid S = s_j) \right|.
    \end{equation}
    \item The \emph{Average Absolute Pairwise Difference}, computing the average disparity across all pairs:
    \begin{equation}
    \label{eq:dp_avg_pair}
    \operatorname{DPD} = \frac{1}{|\mathcal{S}|^2} \sum_{s_i \in \mathcal{S}} \sum_{s_j \in \mathcal{S}} \left| \mathbb{P}(\hat{Y} = 1 \mid S = s_i) - \mathbb{P}(\hat{Y} = 1 \mid S = s_j) \right|.
    \end{equation}
\end{itemize}

Since achieving perfect demographic parity (where all the above differences are zero) can be impractical, the objective is often relaxed to minimizing the Demographic Parity Distance:

\begin{align}
\label{eq:delta_dp_sum}
\Delta^{\text{DP}} =\sum_{s \in \mathcal{S}} \left| \mathbb{P}(\hat{Y} = 1 \mid S=s) - \mathbb{P}(\hat{Y} = 1 )\right|
.
\end{align}

While demographic parity is often defined in terms of these outcome rate differences, it can also be understood more generally in terms of statistical distance between group-conditional distributions of model predictions or representations \cite{madras2018learning}. Let $Z_0$ and $Z_1$ denote the distributions of these representations conditioned on different sensitive attribute values $S=0$ and $S=1$, respectively. Then, for any measurable test function $f_\theta$ applied to the model outputs, one can define the test discrepancy as:
\begin{equation}
\Delta^{\text{DP}}(f_\theta \circ h_\theta) = \left\| \mathbb{E}_{Z_0}[f_\theta(Z)] - \mathbb{E}_{Z_1}[f_\theta(Z)] \right\|.
\end{equation}
The statistical distance (or total variation distance) between $Z_0$ and $Z_1$ is the supremum of this discrepancy over all measurable test functions $\mu$:
\begin{equation}
\Delta^*(Z_0, Z_1) = \sup_{\mu} \left\| \mathbb{E}_{Z_0}[\mu(Z)] - \mathbb{E}_{Z_1}[\mu(Z)] \right\|.
\end{equation}
It follows that for any $f_\theta$,
\begin{equation}
\Delta^{\text{DP}}(f_\theta \circ h_\theta) \leq \Delta^*(Z_0, Z_1),
\end{equation}
and $\Delta^{\text{DP}}(f_\theta \circ h_\theta) = 0 $ if and only if $f_\theta(h_\theta(X)) \perp S $, i.e., demographic parity holds.

In contrast to Demographic Parity, the Equal Opportunity (EO) criterion incorporates information about the target variable $Y$ \cite{hardt2016equality}. Specifically, EO requires that the true positive rate (TPR) be equal across all demographic groups. Formally:
\begin{equation}
\label{eq:eop_ideal}
\mathbb{P}(\hat{Y} = 1 \mid S=s, Y=1) = \mathbb{P}(\hat{Y} = 1 \mid Y=1) \quad \forall s \in \mathcal{S}
.
\end{equation}
This fairness criterion is especially salient in domains where false negatives are particularly harmful and reasonably accurate ground truth labels are available, such as in healthcare, criminal justice, and risk assessment \cite{chouldechova2017fair, seyyedkalantari2020chexclusionfairnessgapsdeep, obermeyer2019dissecting}. 
The corresponding EO distance is defined as:
\begin{equation}
    \Delta^{\text{EO}}(f_{\theta} \circ h_{\theta}) = \left\| \mathbb{E}_{Z}[f_{\theta}(Z)\mid S=0, Y=1] - \mathbb{E}_{Z}[f_{\theta}(Z)\mid S=1, Y=1] \right\|
\end{equation}
which can be upper bounded by the objective value of an optimal adversary $g^*$ trained to distinguish between $Z \mid Y=1$ for different groups \cite{madras2018learning}.


An alternative fairness criterion is Equalized Odds (EOd) \cite{hardt2016equality}, which extends Equal Opportunity by requiring that prediction outcomes be conditionally independent of the sensitive attribute given the true label. Formally, EOd demands that for all $s \in \mathcal{S}$:
\begin{align}
    \measure{}(\hat{Y}=1 \mid Y=1, S=s) &= \measure{}(\hat{Y}=1 \mid Y=1) \quad \forall s \in S
\end{align}
and
\begin{align}
    \measure{}(\hat{Y}=1 \mid Y=0, S=s) &= \measure{}(\hat{Y}=1 \mid Y=0) \quad \forall s \in S
\end{align}
These constraints ensure that both the true positive rate (TPR) and false positive rate (FPR) are equal across all groups \cite{zhong2024intrinsic, golz2019paradoxes}. This criterion is particularly applicable in decision-making contexts where reliable ground truth labels are available and both false negatives and false positives incur significant societal or personal costs \cite{davrazos2025evaluting, Yu_2024}.
The EOd distance is defined as:
\begin{align}
    \Delta^{\text{EOd}}(f_{\theta} \circ h_{\theta}) = & \left\| \mathbb{E}_{Z}[f_{\theta}(Z)\mid S=0, Y=1] - \mathbb{E}_{Z}[f_{\theta}(Z)\mid S=1, Y=1] \right\| \notag\\
    + & \left\| \mathbb{E}_{Z}[f_{\theta}(Z)\mid S=0, Y=0] - \mathbb{E}_{Z}[f_{\theta}(Z)\mid S=1, Y=0] \right\|.
\end{align}

\subsection{Additional empirical evaluations and ablations}
\label{appendix:other_results}
In \cref{tab:adult_other} we compare our proposed approaches to other baselines on the Adult dataset, with the same setup as in \cite{moyer2018invariant}. Given that no distinction is made on the family of functions used for the evaluation, we report the Adv. Accuracy of FmCF and FmSS approaches using a deep and wide \ac{MLP}. The same holds true for the results reported in \cref{tab:german_other}, where we compare out methods to the same setup from \cite{tucker2022prototype}

\begin{table}[h!]
  \centering
  \begin{subtable}[b]{0.45\textwidth}
    \centering
    \caption{Comparison on Adult dataset using the setup from \cite{moyer2018invariant}.}
    \label{tab:adult_other}
    \begin{tabular}{ccc}
      \toprule
      Model & Acc. & Adv. Acc. \\
      \midrule
      Random & 75.2 &  67.5 \\
      VFAE~\cite{louizos2017variationalfairautoencoder}  & 84.2 &  88.2 \\
      CIAFL~\cite{xie2017controllable} & 83.1 & 88.8 \\
      IRwAL~\cite{moyer2018invariant} & 84.2 & 77.6 \\
      \midrule
      FmCF (ours) & 85.0 & 67.8 \\
      FmSS (ours) & 85.0 & 67.9 \\
      \bottomrule
    \end{tabular}
  \end{subtable}
  \hfill
  \begin{subtable}[b]{0.45\textwidth}
    \centering
    \caption{Comparison on Adult dataset using the setup from \cite{tucker2022prototype}.}
    \label{tab:german_other}
    \begin{tabular}{ccc}
      \toprule
      Model & Acc. & Adv. Acc. \\
      \midrule
      Random & 70.0 & 81.0 \\
      CSN~\cite{tucker2022prototype} & 73.1 & 81.3 \\
      CIAFL~\cite{xie2017controllable} & 73.6 & 81.1 \\
      VFAE~\cite{louizos2017variationalfairautoencoder} & 72.8 & 81.2 \\
      FRTrain~\cite{roh2020fr} & 72.7 & 80.9 \\
      WassDB~\cite{jiang2020wasserstein} & 72.8 & 81.1 \\
      \midrule
      FmCF (ours) & 74.1 & 81.1 \\
      FmSS (ours) & 74.2 & 81.1 \\
      \bottomrule
    \end{tabular}
  \end{subtable}
\end{table}

\begin{figure}[H]
    \centering
    \includegraphics[width=\linewidth]{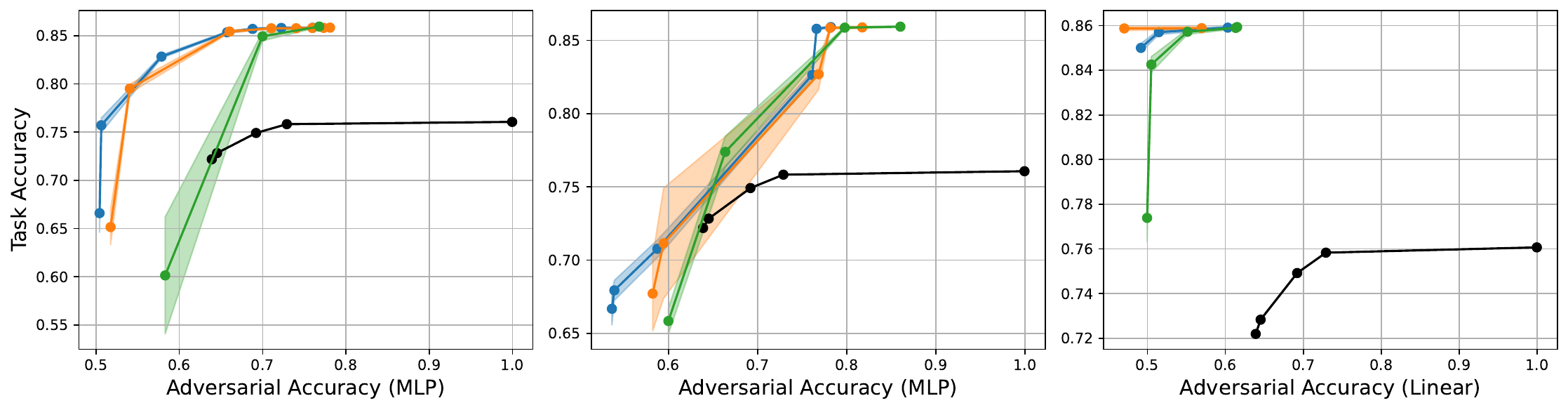}
    \caption{Pareto varying latent dimension on Health dataset.}
    \label{fig:pareto_healt}
\end{figure}

\begin{figure}[H]
    \centering
    \includegraphics[width=\linewidth]{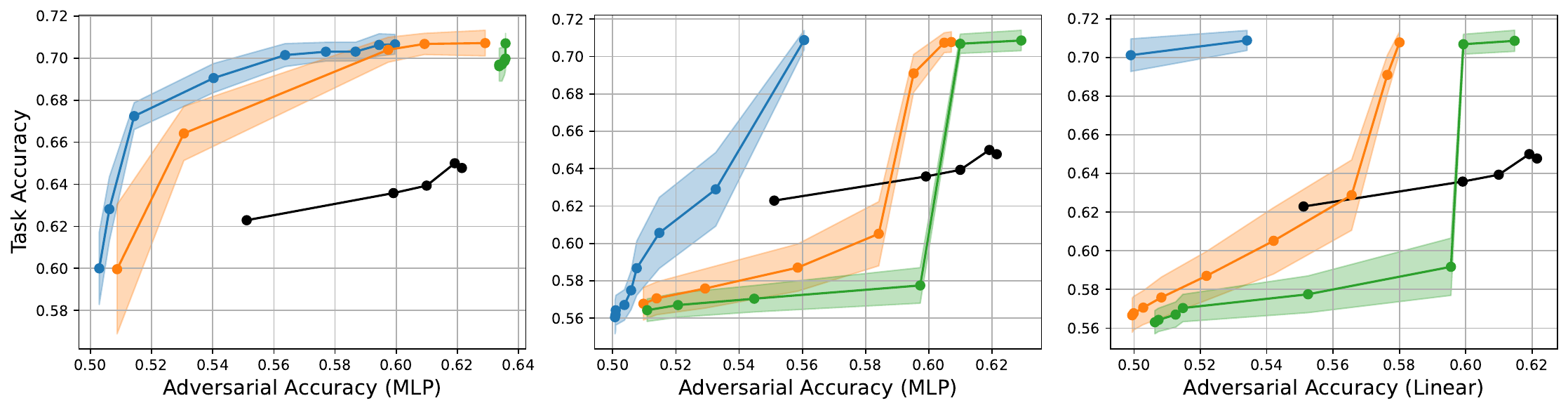}
    \caption{Pareto varying latent dimension on Compas dataset.}
    \label{fig:pareto_compas}
\end{figure}

\begin{figure}[H]
    \centering
    \includegraphics[width=\linewidth]{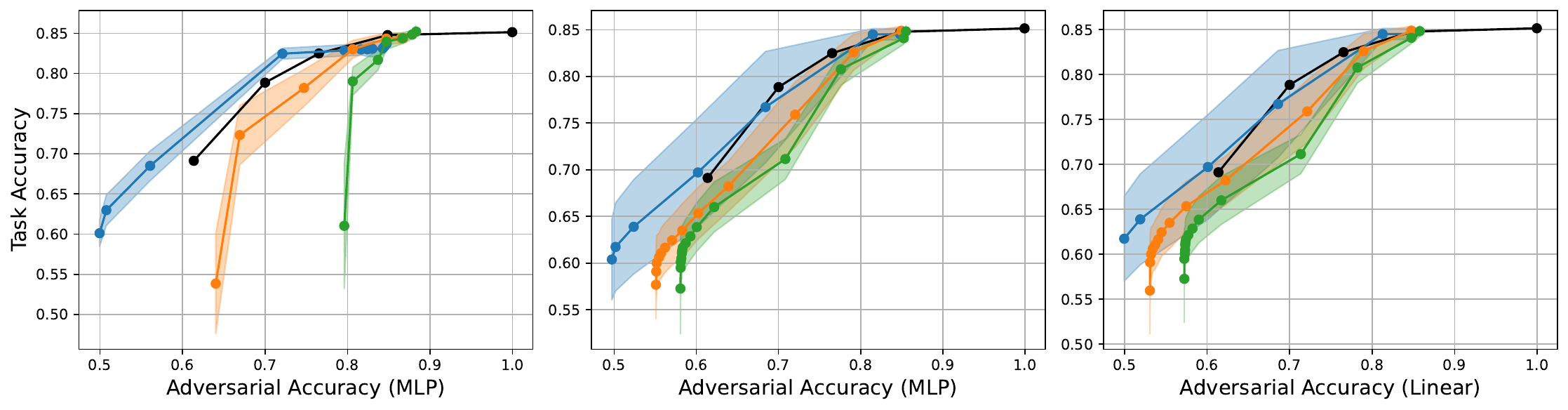}
    \caption{Pareto varying latent dimension on Crime dataset.}
    \label{fig:pareto_crime}
\end{figure}

\begin{figure}[H]
    \centering
    \includegraphics[width=\linewidth]{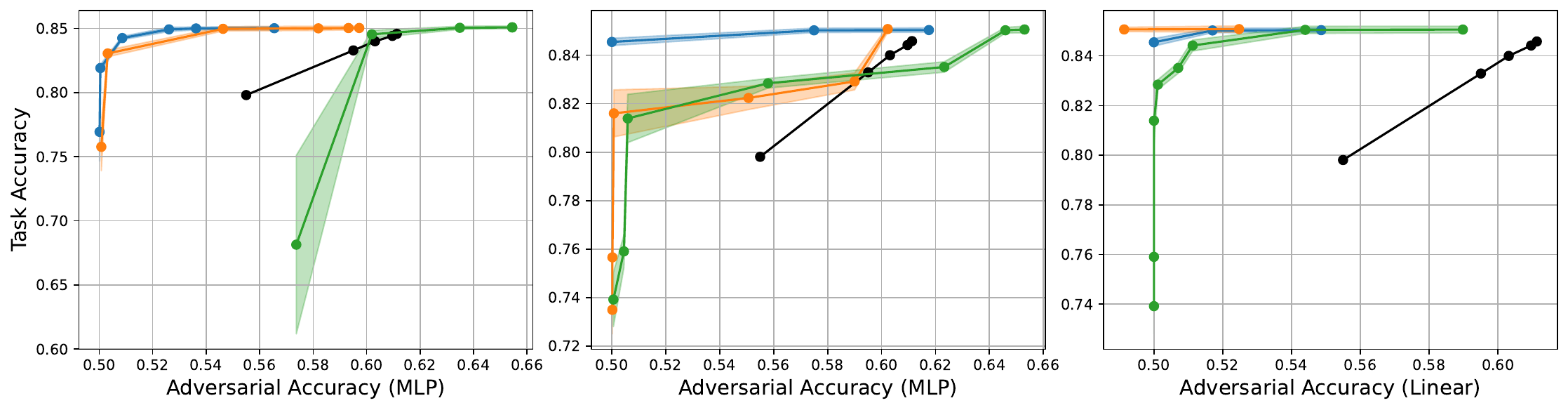}
    \caption{Pareto varying latent dimension on Adult dataset.}
    \label{fig:pareto_adult}
\end{figure}

In \cref{fig:pareto_adult,fig:pareto_compas,fig:pareto_crime,fig:pareto_healt}, we report the tradeoffs between adversarial balanced accuracy and task accuracy. We evaluate FmCF \cref{sec:fair_CF} against an adversarial \ac{MLP} as reported in \cref{appendix:training_details}. 
Instead, we evaluate FmSS \cref{sec:fair_SS} both against an \ac{MLP} and against a \ac{LR}. 
In particular, for all images, in black is reported the performance of Fair Normalizing Flows \cite{balunovic2022fair}, while in blue is reported the performance using 1 dimension, in orange 2 dimensions, and in green 3 dimensions. We observe that the lower the dimension, the easier it is for the penalties to be effective, giving surprising performances for extremely low-dimensional latent spaces. Notably, such a characteristic is not limiting for the downstream task, as even in the lowest-dimensional setting, both approaches outperform all the baselines.

Furthermore, we evaluated the effects of batch size on the reliability of the CFD estimation. As confirmed by \cref{tab:estimation_variance_transposed}, larger batches yield more stable and reliable CFD estimates. The variance of the estimates decreases consistently with increasing batch size, aligning with the expected theoretical scaling of 
$O(\frac{1}{n})$.

\begin{table}[h!]
\label{tab:estimation_variance_transposed}
\centering
\caption{Standard deviation of FmCF Penalty estimator.}
\begin{tabular}{lcccc}
\hline
\textbf{Batch Size} & \textbf{Adult} & \textbf{German} & \textbf{Compas} & \textbf{Health} \\
\hline
8   & 0.1150 & 0.1252 & 0.0924 & 0.1004 \\
16  & 0.0596 & 0.0854 & 0.0361 & 0.0422 \\
32  & 0.0164 & 0.0172 & 0.0129 & 0.0121 \\
64  & 0.0063 & 0.0056 & 0.0052 & 0.0054 \\
128 & 0.0026 & 0.0024 & 0.0022 & 0.0027 \\
256 & 0.0016 & 0.0004 & 0.0010 & 0.0015 \\
512 & 0.0007 & 0.0000 & 0.0007 & 0.0009 \\
\hline
\end{tabular}
\end{table}

\begin{figure}[h]
    \centering
    \includegraphics[width=0.8\linewidth]{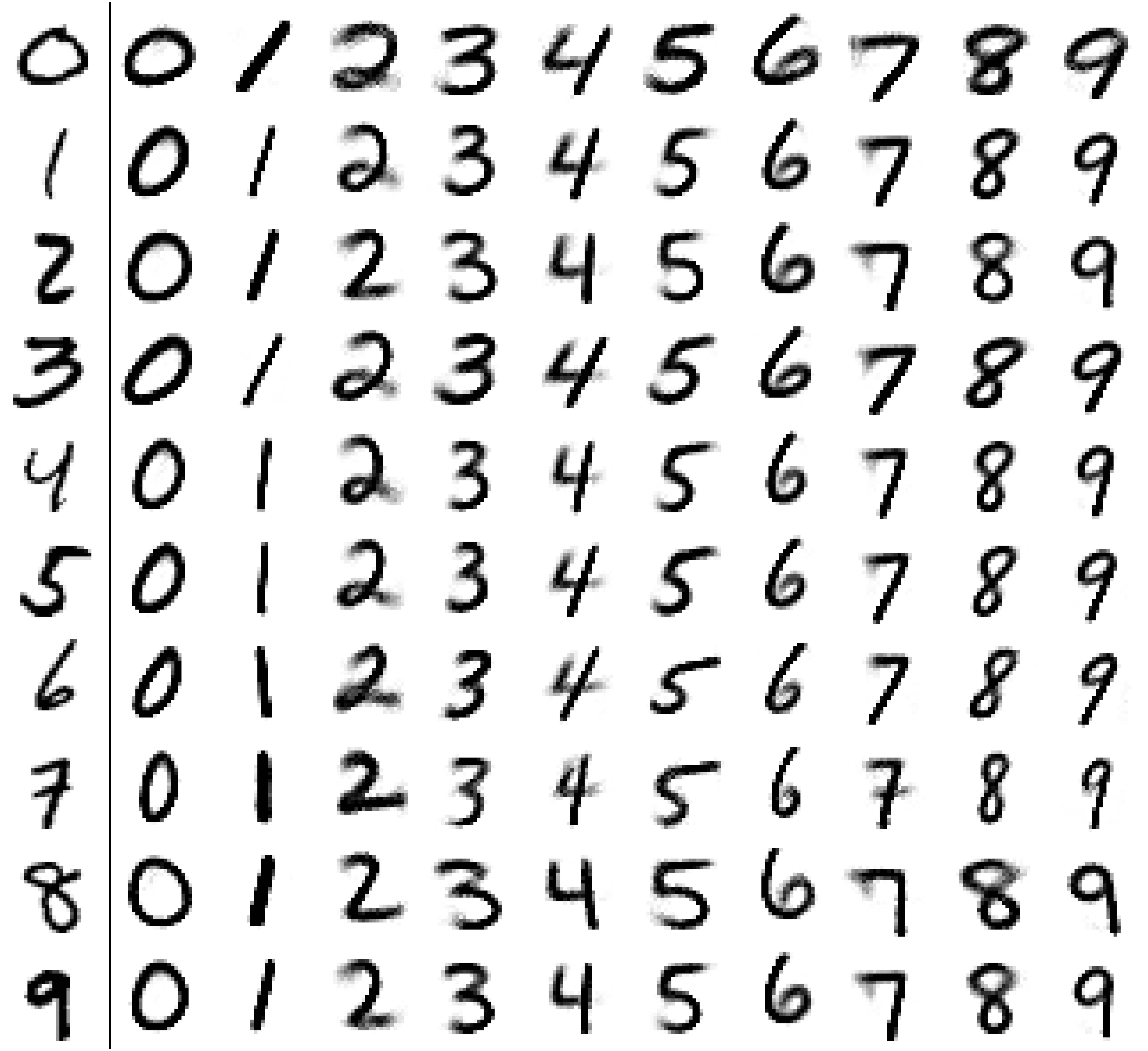}
    \caption{Latent representations $Z$ are extracted from the left-most column and combined with arbitrary target labels to generate stylistically consistent images of different classes, shown on the right.}
    \label{fig:mnist}
\end{figure}

Given that all benchmarks considered thus far pertain to classification tasks, we now present a simple example demonstrating the applicability of FmCF to alternative problem settings. Specifically, in \cref{fig:mnist}, a conditional convolutional autoencoder is employed to learn a latent representation of the MNIST handwritten digits dataset. Subsequently, the loss function described in \cref{alg:fmcf_short} is utilized to penalize the divergence between $\measure{Z \mid S}$ and a Gaussian distribution. When this condition is satisfied, it ensures that $Z \perp\!\!\!\perp S$, indicating that the latent representation $Z$ encodes information about $X$ that is independent of $S$. Consequently, when the model is prompted to reconstruct an input with a different label, it retains the essential characteristics of the original input while generating a digit consistent with the newly specified label. 

Indeed, the decoder is able to generate images preserving characteristics such as rotation, while only relying on the fed label for the kind of digit to show.

For more details on the datasets, please refer to \cref{appendix:datasets}.

\subsection{Datasets}
\label{appendix:datasets}
We utilize six well-known datasets in our study, sourced from both the UCI Machine Learning Repository and other publicly available resources. These datasets include Adult, Crime, Compas, Law School, Health, and the Statlog (German Credit Data) dataset. Each of these datasets presents distinct challenges and characteristics relevant to fairness and predictive modeling. Below, we briefly introduce each dataset and outline the preprocessing steps employed:
\begin{itemize}
    \item Adult\footnote{\url{https://archive.ics.uci.edu/dataset/2/adult}}: The Adult dataset, also known as the Census Income dataset, originates from the 1994 Census database and is available through the UCI repository~\cite{dua2019uci}. It contains 14 attributes, including age, workclass, education, marital status, occupation, race, and sex. The prediction task is to determine whether an individual’s income exceeds \$50,000 per year. To facilitate modeling, we discretize continuous features and retain categorical variables related to demographics and employment. Sex is treated as the protected attribute in fairness analyses.
    
    \item Crime\footnote{\url{https://archive.ics.uci.edu/dataset/183/communities+and+crime}}: The Communities and Crime dataset combines socio-economic data from the 1990 US Census, law enforcement data from the 1990 LEMAS survey, and crime data from the 1995 FBI UCR. The goal is to predict whether the violent crime rate of a community is above or below the median. We utilize attributes such as race percentages, income levels, and family structure indicators. Race is designated as the protected attribute, derived from the proportions of racial groups within each community.
    
    \item Compas\footnote{\url{https://github.com/propublica/compas-analysis}}: The Compas dataset contains data related to criminal history, jail and prison time, demographics, and COMPAS risk scores from Broward County (2012–2013)~\cite{brennan2009compas}. The prediction task involves forecasting recidivism within two years. Key attributes include age, prior count, and charge degree, with race being the protected attribute. We preprocess the dataset by discretizing continuous variables and retaining categorical ones, critical for risk prediction.
    
    \item Health\footnote{\url{https://paperswithcode.com/dataset/heritage-health-prize}}: The Health dataset was part of the Heritage Health Prize competition on Kaggle and contains medical records of over 55,000 patients. We focus on the merged claims, drug count, and lab count attributes while removing personal identifiers to ensure privacy. Age is treated as a protected attribute, divided into binary groups: above and below 60 years. The primary prediction task is to assess the maximum Charlson Comorbidity Index, reflecting the long-term survival prospects of patients with multiple conditions.
    
    \item Law School\footnote{\url{https://eric.ed.gov/?id=ED469370}}: The Law School dataset consists of data from admissions cycles between 2005 and 2007, covering over 100,000 individual applications. Attributes include LSAT scores, undergraduate GPA, race, gender, and residency status. To enhance privacy, data has been aggregated where necessary. We consider race as the protected attribute, binarizing it into white and non-white categories. The main task is to predict law school admission outcomes.
    
    \item German\footnote{\url{https://archive.ics.uci.edu/dataset/522/south+german+credit}}: The German Credit Data dataset, sourced from the UCI Machine Learning Repository~\cite{dua2019uci}, consists of 1,000 records of credit applicants. Each instance is labeled as either "good" or "bad" credit risk. The dataset contains 20 features, including age, credit amount, employment status, and personal status, among others. Both categorical and numerical data are present, requiring careful preprocessing. We encode categorical variables using one-hot encoding and normalize numerical features. The protected attribute for fairness evaluation in this dataset is age, segmented into groups representing different age ranges.
\end{itemize}

However, in \cite{balunovic2022fair}, the Lawschood dataset is also used to show the effectiveness of FNF. Through the extensive research, we did not find any publicly available version of such a dataset, thus our approaches are not evaluated on it.

\subsection{Training details}
\label{appendix:training_details}
All experiments reported in this paper were implemented using PyTorch. The models were trained on a server equipped with an AMD Ryzen Threadripper PRO 5995WX CPU (64 cores, 128 threads), 512 GiB of RAM, and three NVIDIA RTX A6000 GPUs. Despite this powerful setup, none of the models utilized more than 4 GiB of VRAM during training.

All reported results are averages over 10 different seeds of the proposed approaches. No extensive hyperparameter tuning was performed for any of the reported results, underscoring the effectiveness of the proposed methods. All \ac{MLP}s used for adversarial evaluations, encoding, and classification consist of four layers with 64 neurons each. Additionally, larger \ac{MLP}s for fairness adversarial evaluations, configured with 128 and 256 neurons, were tested but resulted in poorer performance.

The Adam optimizer \cite{kingma2014adam} was employed for all training sessions, with a learning rate of 0.0003. Training was conducted for 100 epochs, incorporating L2 regularization with a weight penalty of 0.0001 to mitigate overfitting. Since accuracy and fairness often involve a trade-off, early stopping was not applied. Instead, the model from the final epoch was used for evaluation.

\newpage
\section*{NeurIPS Paper Checklist}
\begin{enumerate}
	
	\item {\bf Claims}
	\item[] Question: Do the main claims made in the abstract and introduction accurately reflect the paper's contributions and scope?
	\item[] Answer: \answerYes{} 
	\item[] Justification: In the abstract, we introduce prior works highlighting limitations in approaches to learning specialized representations. We then show theoretically and empirically that the proposed approaches learn specialized representations without any adversarial loss in an effective manner, surpassing the current state-of-the-art.
	
	\item {\bf Limitations}
	\item[] Question: Does the paper discuss the limitations of the work performed by the authors?
	\item[] Answer: \answerYes{} 
	\item[] Justification: In the conclusions (and briefly in the sections presenting the new methods), limitations are highlighted, bringing attention to the difficulty of learning high-dimensional fair representations. 
	
	\item {\bf Theory assumptions and proofs}
	\item[] Question: For each theoretical result, does the paper provide the full set of assumptions and a complete (and correct) proof?
	\item[] Answer: \answerYes{} 
	\item[] Justification: In the appendix, we provide formal proofs for each theorem in the main text, with numbered equations. In the main text, we provide visual examples and theorem statements.
	
	\item {\bf Experimental result reproducibility}
	\item[] Question: Does the paper fully disclose all the information needed to reproduce the main experimental results of the paper to the extent that it affects the main claims and/or conclusions of the paper (regardless of whether the code and data are provided or not)?
	\item[] Answer: \answerYes{} 
	\item[] Justification: In the appendix, we provide information on the used datasets, preprocessing, and training, with details on the setup and hardware being used. Furthermore, for easy better interpretation, we include in the main text a pseudocode of the proposed approaches. Finally, we provide the code as supplementary material.

	\item {\bf Open access to data and code}
	\item[] Question: Does the paper provide open access to the data and code, with sufficient instructions to faithfully reproduce the main experimental results, as described in supplemental material?
	\item[] Answer: \answerYes{} 
	\item[] Justification: Datasets are publicly available. We provide the code, preprocessing, configurations, and dataloaders to reproduce all results in the paper.

	\item {\bf Experimental setting/details}
	\item[] Question: Does the paper specify all the training and test details (e.g., data splits, hyperparameters, how they were chosen, type of optimizer, etc.) necessary to understand the results?
	\item[] Answer: \answerYes{} 
	\item[] Justification: In the appendix, we address all the details of the training. Furthermore, the provided code is already set to the same parameters used for creating such results. No hyperparameter tuning has been carried out for the proposed methods.
	
	\item {\bf Experiment statistical significance}
	\item[] Question: Does the paper report error bars suitably and correctly defined or other appropriate information about the statistical significance of the experiments?
	\item[] Answer: \answerYes{} 
	\item[] Justification: We provide error bars in all the Pareto plots. However, given that all other approaches did not report the error bars, we do not report them in the tables.
	
	\item {\bf Experiments compute resources}
	\item[] Question: For each experiment, does the paper provide sufficient information on the computer resources (type of compute workers, memory, time of execution) needed to reproduce the experiments?
	\item[] Answer: \answerYes{} 
	\item[] Justification: In the appendix, all hardware specifications and usages are specified.
	
	\item {\bf Code of ethics}
	\item[] Question: Does the research conducted in the paper conform, in every respect, with the NeurIPS Code of Ethics \url{https://neurips.cc/public/EthicsGuidelines}?
	\item[] Answer: \answerYes{} 
	\item[] Justification: We include reproducibility details, the data are publicly available, and we address the broader impact and societal implications at the end of the main paper.

	\item {\bf Broader impacts}
	\item[] Question: Does the paper discuss both potential positive societal impacts and negative societal impacts of the work performed?
	\item[] Answer: \answerYes{} 
	\item[] Justification: Though no major broader impact can be foreseen, we partially address such concerns at the end of the main paper. However, the whole paper aims at developing fair classifiers which has an inherent positive societal impact.
	
	\item {\bf Safeguards}
	\item[] Question: Does the paper describe safeguards that have been put in place for responsible release of data or models that have a high risk for misuse (e.g., pretrained language models, image generators, or scraped datasets)?
	\item[] Answer: \answerNA{} 
	\item[] Justification: We find that this paper poses no such risk.
	
	\item {\bf Licenses for existing assets}
	\item[] Question: Are the creators or original owners of assets (e.g., code, data, models), used in the paper, properly credited and are the license and terms of use explicitly mentioned and properly respected?
	\item[] Answer: \answerYes{} 
	\item[] Justification: We work with open source libraries and datasets that are publicly available and we cited them properly.
	
	\item {\bf New assets}
	\item[] Question: Are new assets introduced in the paper well documented and is the documentation provided alongside the assets?
	\item[] Answer: \answerYes{} 
	\item[] Justification: We provide code and instructions on how to run the code in order to reproduce our results.
	
	\item {\bf Crowdsourcing and research with human subjects}
	\item[] Question: For crowdsourcing experiments and research with human subjects, does the paper include the full text of instructions given to participants and screenshots, if applicable, as well as details about compensation (if any)? 
	\item[] Answer: \answerNA{} 
	\item[] Justification: The paper does not involve crowdsourcing nor research with human subjects
	
	\item {\bf Institutional review board (IRB) approvals or equivalent for research with human subjects}
	\item[] Question: Does the paper describe potential risks incurred by study participants, whether such risks were disclosed to the subjects, and whether Institutional Review Board (IRB) approvals (or an equivalent approval/review based on the requirements of your country or institution) were obtained?
	\item[] Answer: \answerNA{} 
	\item[] Justification: The paper does not involve crowdsourcing nor research with human subjects
	
	\item {\bf Declaration of LLM usage}
	\item[] Question: Does the paper describe the usage of LLMs if it is an important, original, or non-standard component of the core methods in this research? Note that if the LLM is used only for writing, editing, or formatting purposes and does not impact the core methodology, scientific rigorousness, or originality of the research, declaration is not required.
	\item[] Answer: \answerNo{} 
	\item[] Justification: LLMs have been used only for paraphrasing and grammar checking.
	
\end{enumerate}

\end{document}